\documentclass[10pt,twocolumn,letterpaper]{article}

\usepackage{iccv}              
\usepackage{multirow}
\usepackage{arydshln} 
\usepackage{graphicx}    
\usepackage{pifont} 
\usepackage[table]{xcolor} 
\usepackage{colortbl}  
\usepackage[table]{xcolor}
\usepackage{algorithm}
\usepackage{algorithmicx}
\usepackage{algpseudocode}
\usepackage[accsupp]{axessibility}  

\newcolumntype{?}{!{\color{gray!30}\vrule}}

\newcommand{\cmark}{\ding{51}}%
\newcommand{\xmark}{\ding{55}}%
\definecolor{iccvblue}{rgb}{0.21,0.49,0.74}
\usepackage[pagebackref,breaklinks,colorlinks,allcolors=iccvblue]{hyperref}

\def\paperID{6686} 
\def\confName{ICCV}
\def\confYear{2025}

\title{ProbMED: A Probabilistic Framework for Medical Multimodal Binding}

\author{
Yuan Gao\textsuperscript{1,2,3,5,6}\thanks{Equal contribution to this work.}
\quad
Sangwook Kim\textsuperscript{1,3,4,5,6}\footnotemark[1]
\quad
Jianzhong You\textsuperscript{1,2,3,5,6}
\quad
Chris McIntosh\textsuperscript{1,2,3,4,5,6}
\\[1ex]
\textsuperscript{1}Peter Munk Cardiac Centre \quad
\textsuperscript{2}Ted Rogers Centre for Heart Research \quad 
\textsuperscript{3}University Health Network \quad\\
\textsuperscript{4}Joint Department of Medical Imaging \quad 
\textsuperscript{5}University of Toronto \quad
\textsuperscript{6}Vector Institute \\
{\tt\small \{yuan.gao, sangwook.kim, jianzhong.you, chris.mcintosh\}@uhn.ca}
}
\def\thickhline{\noalign{\hrule height1pt}}
\begin{document}
\maketitle
\begin{abstract}
Medical decision-making requires integrating diverse medical information, from imaging to clinical narratives. These medical modalities are often acquired in a many-to-many manner.  However, current medical vision-language pretraining models (Med-VLPMs) fail to directly account for this many-to-many mapping in their model training and embeddings. To address this, we present \textbf{P}robabilistic \textbf{M}odality-\textbf{E}nhanced \textbf{D}iagnosis (ProbMED), a multimodal Med-VLPM that employs probabilistic contrastive learning to model distributions over embeddings rather than deterministic estimates. ProbMED aligns four distinct modalities—chest X-rays, electrocardiograms, echocardiograms, and clinical text—into a unified probabilistic embedding space. We use InfoNCE loss with Hellinger distance to integrate inter-modality distributions. We introduce a probabilistic synthetic sampling loss that captures modality-specific mean and variance to improve intra-modality binding. Extensive experiments across 13 medical datasets demonstrate that our model outperforms current Med-VLPMs in cross-modality retrieval, zero-shot, and few-shot classification. We also demonstrate the robust integration of multiple modalities for prognostication, showing improved intra- and inter-medical modality binding. Code is available:
 \href{https://github.com/mcintoshML/probMED}{https://github.com/mcintoshML/probMED}.
\end{abstract}    
\section{Introduction}
\label{sec:intro}
Medical decision-making is inherently multimodal and requires integrating diverse information ranging from imaging modalities to clinical reports. Despite the growing success of medical vision language pretraining models (Med-VLPM) in extracting embeddings from paired modalities, typically chest radiographs (CXR) with corresponding reports, these approaches operate primarily under a deterministic embedding framework that enforces one-to-one mappings~\cite{gao2024medbind, huang2021gloria, you2023cxrclip, tiu2022expertchexzero, christensen2024visionechoclip,zhang2023biomedclip}. Existing approaches face two key limitations: \textbf{1)} Deterministic models may struggle to capture the inherent variability and complex many-to-many relationships in medical data, \textbf{2)} Majority of existing models focus exclusively on CXR-text alignment, overlooking broader multimodal nature of medical care.

Regarding the \textit{first} problem, there are many medical cases where many-to-many relationships exist. For example, a CXR of a patient in respiratory distress can have multiple valid interpretations: \texttt{"A cloudy patch in the lower lung"} or \texttt{"CXR has pneumonia"}. Although phrased differently, these examples inherently convey the same meaning, where one \textbf{describes} pneumonia and the other \textbf{states} the disease. Thus, these two examples should both relate to a pneumonia CXR, but not all cloudy patches are pneumonia (e.g., cloudy patches could be lung cancer). Next, consider a patient visit; they may require multiple electrocardiograms (ECGs) and CXRs to confirm prognosis--the relationships between ECGs and CXRs are inherently \textbf{many-to-many}. These examples highlight the limitations of deterministic methods that force embeddings into discrete positive/negative labels, making them ill-suited to modeling the ambiguity in the pairings. Recent advances in contrastive learning have highlighted the importance of capturing these relationships in learned representations~\cite{chun2023improvedpcmepp}. Probabilistic contrastive learning extends traditional methods by modeling distributions over embeddings rather than single-point estimates~\cite{chun2021probabilistic, kirchhof2023probabilistic}; thus, each instance is represented as a distribution, which can enable better semantic overlap and \textit{resolve} ambiguity during training. 

For the \textit{second} problem, real-world diagnostics integrate multiple modalities, for a more comprehensive clinical picture~\cite{tu2024towards, gao2024medbind}. However, as modalities multiply, cross-modal pairings grow quadratically, requiring a probabilistic contrastive approach. Thus, our approach unifies multiple medical data pairs through probabilistic contrastive learning by randomly sampling one modality pair per gradient update.

 \begin{figure*}
    \centering
    \includegraphics[width=\textwidth]{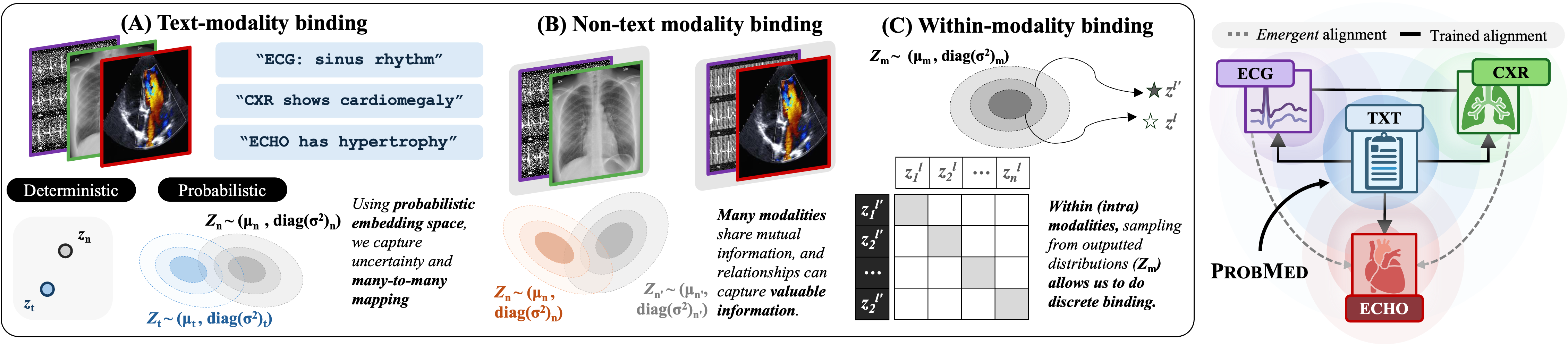}
    \caption{The overview of \textbf{\textsc{ProbMED}}. (Left) \textbf{(A) Text-modality:} Given multiple medical modalities, we align each with its corresponding textual description using a contrastive learning framework. Traditional \textit{deterministic} approaches represent each modality-text pair as fixed points in latent space ($z_n$ and $z_t$), whereas \textit{probabilistic} ones model each modality embedding as a Gaussian distribution ($Z_n$ and $Z_t$), detailed in $\S$\ref{sec:bindingmodal}. \textbf{(B) Non-text modality:} We also bind \textit{between non-text modalities} and model overlapping related modalities, where $n$ and $n'$ are different non-text modalities ($\S$\ref{sec:bindingmodal}). \textbf{(C) Within-modality:} We introduce a Synthetic Instance Sampling Loss for improved \textit{within}-modality binding. Given any modality $Z_m$, we use the learned distribution to sample additional samples $z^{l}$ and $z^{l'}$, detailed in $\S$\ref{sec:intramodal}. (Right) \textbf{\textsc{ProbMED}} links ECG, CXR, ECHO, and text into a unified probabilistic embedding space, we trained (\textit{\textbf{black solid lines}}) on all text-modality pairs and ECG-CXR (non-text modality) pairs. We observe \textit{emergent} alignment (\textcolor{Gray}{\textbf{\textit{grey dashed lines}}}) after training our model.}
    \label{fig:framework}
\end{figure*}

In this work, we introduce \textbf{Prob}abilistic \textbf{M}odality-\textbf{E}nhanced \textbf{D}iagnosis (\textsc{\textsc{\textsc{ProbMED}}}), a probabilistic multimodal Med-VLPM that bridges the gap between multiple medical modalities. This is the first study to leverage probabilistic modeling for such extensive multimodal integration in the medical domain. Unlike prior approaches~\cite{chun2023improvedpcmepp,chun2021probabilisticpcmeog}, \textsc{\textsc{\textsc{ProbMED}}} uses a probabilistic contrastive framework to cross-integrate \textit{four} distinct medical modalities: CXR, ECG, echocardiogram (ECHO), and text.

\noindent\textbf{Contributions:} \textbf{1)} We propose \textbf{\textsc{ProbMED}}, which integrates multimodal medical data through probabilistic mappings. It aligns learned probability distributions across modalities within a shared embedding space while leveraging a novel loss function, Synthetic Instance Sampling Loss, to enhance intra-modal representation. \textbf{2)} We evaluated \textsc{\textsc{ProbMED}} on 13 distinct medical datasets, demonstrating superior performance in cross-modality retrieval, zero-shot, and few-shot classification. \textbf{3)} We further showed its multimodal capability by integrating CXR and ECG for improved prognostication of diseases often misdiagnosed using a single modality~\cite{pan2021prognosticCKDCXR}. \textbf{4)} We showcased use-cases of the probabilistic modeling enabled by \textsc{ProbMED}: \textit{Uncertainty-based prompt filtering} to enhance robustness against ambiguous data pairs, and \textit{Distribution-based sampling} to improve classification in few-shot scenarios. 

\section{Related Work}
\noindent\textbf{Contrastive Learning.}
We focus on cross-modal contrastive learning, which integrates multimodal data like images and text~\cite{radford2021learning_clip}. Cross-modal learning has proven especially valuable in the medical domain~\cite{you2023cxrclip, wang2022medclip, zhang2023biomedclip, tiu2022expertchexzero, christensen2024visionechoclip, zhang2022contrastiveconvirt}. The objective is to maximize the similarity between modality pairs, e.g., aligning a CXR with its corresponding radiology report. These foundational approaches generate semantically rich representations that can be fine-tuned for downstream tasks, such as disease classification, even with limited annotated datasets. The ability to learn from data-scarce scenarios is a significant focus of this study since it is especially valuable in medical research, where acquiring large-scale annotated datasets is limited.

\noindent\textbf{Probabilistic Multimodal Embeddings.}
Probabilistic contrastive learning enhances cross-modal learning in natural images by explicitly modeling the ambiguity in mapping visual features to textual descriptions~\cite{chun2021probabilistic, upadhyay2023probvlm, chun2023improvedpcmepp, kirchhof2023probabilistic}. This ambiguity arises primarily from the prevalence of false-negative pairs during training~\cite{chun2021probabilisticpcmeog}.  Early methods, such as Probabilistic Cross-Modal Embedding (PCME)~\cite{chun2021probabilistic}, introduced probabilistic embeddings to move beyond fixed representations but suffered from high computational costs and loss saturation. PCME++~\cite{chun2023improvedpcmepp} mitigates these challenges by introducing a closed-form probabilistic distance (CSD) and pseudo-positive samples with mixed-sample augmentations, enhancing learning of many-to-many relationships. 

As described, the many-to-many challenge is also evident in the medical domain. Probabilistic embeddings can thus offer a dual benefit: they enable flexible cross-modal correspondences and explicitly capture uncertainty (via mean and variance), leading to richer representations for downstream tasks in data-scarce scenarios. Extending these models to bind more than three or more modalities remains an open challenge--one we address here.

\noindent\textbf{Multimodal Learning.} 
Multimodal learning extends many learning principles beyond simple image-text pairs to integrate additional modalities.  ImageBind~\cite{girdhar2023imagebind} introduced an effective way to integrate multiple modality pairs into a single unified model, improving the understanding of cross-modal mappings. Although the study was conducted on natural images, the potential of multimodal training in the medical domain is significant~\cite{moor2023foundation,tu2024towards,saab2024capabilitiesmedgemini,poon2023multimodal, xiang2025visionMUSK}. For instance, the use of medical data such as computed tomography~\cite{hamamci2024developingctclip,You2025X2CT}, sensor signals~\cite{gao2024medbind,mckeen2024ecg}, endoscopic videos~\cite{boers2024foundationendo}, and even genomic information~\cite{ye2024genomics} are emerging. As such, it can be seen that the multi-pair training paradigm capitalizes on each modality's complementary strengths, allowing for models that generalize across diverse clinical scenarios and tasks, especially for disease assessments that require examination of multiple modalities~\cite{pan2021prognosticCKDCXR}.
\section{Methods}
In this section, we introduce \textsc{ProbMED}, which learns a unified joint probabilistic embedding space for multiple medical modalities by utilizing all possible data pairs, with clinical notes to connect them (functioning as the primary binder). Here, each modality’s embeddings are aligned with their corresponding text embeddings (e.g., CXR to radiology text) and/or across modalities (e.g., CXR to its corresponding ECG) as shown in~\cref{fig:framework}. Inspired by ImageBind~\cite{girdhar2023imagebind}, \textsc{\textsc{ProbMED}} does not require complete pairs of modalities (that is, the four modalities of a single patient) during training, making it more practical for real-world datasets. We hypothesize that the resulting probabilistic embedding space across multiple modalities can better deal with the many-to-many mapping typically found in medical data. For the rest of this section, we explain how we trained \textsc{ProbMED} to integrate multiple medical modalities using a probabilistic approach.

\subsection{Aligning Specific Pairs of Data.}
\label{subsection:preliminaries}
Probabilistic contrastive learning is a technique for representing embeddings as distributions in an embedding space. Like traditional contrastive learning, the premise is based on using pairs of related examples (positives) and unrelated examples (negatives) to align data optimally. The key difference lies in the representation: instead of a fixed-point deterministic representation, the feature extractor outputs a distribution parameterized by mean, $\mu$, and covariance matrix, $\boldsymbol{\Sigma}$. As proposed in ~\cite{chun2023improvedpcmepp}, we stipulate that the covariance matrix is strictly the \textbf{diagonal covariance} matrix: 
\begin{equation}
\label{eq:premise}
    \boldsymbol{\Sigma} = \operatorname{diag}(\sigma^2),
\end{equation}
where $\sigma^2$ represents the variance in each dimension. As such, we define our distributions as:
\begin{equation}
\label{eq:defined_distribution}
    Z \sim \mathcal{N}(\mu,\, \operatorname{diag}(\sigma^2)),
\end{equation}
where $\mu,\ \sigma^2 \in \mathbb{R}^D$ for a $D$-dimensional embedding. The pair similarities are computed using probabilistic similarities ($\S$\ref{sec:bindingmodal}), which optimize the joint embedding of different data pairs represented as distributions, allowing for greater flexibility to address many-to-many relationships better.

\subsection{Model Framework}
\textsc{ProbMED} processes four medical modalities—CXR, ECG, ECHO, and medical text—using dedicated modality-specific encoders inspired by~\cite{girdhar2023imagebind}. Supplementary~\cref{sfig:modelarchitecture} presents the architecture overview; however, by design, the model framework is simple to implement. As in~\cite{chun2023improvedpcmepp}, our model relies on pretrained weights to facilitate the transition from deterministic to a probabilistic embedding space.  Specifically, for the CXR encoder, we employ Swin Transformer backbone pretrained with ImageNet-1k~\cite{liu2021swin, huggingface2023swintiny}. For the ECG encoder, we adopt an XResNet-1d-101, a widely used ECG encoder~\cite{strodthoff2020deepxresnet}. For the text encoder, we use BioBERT~\cite{lee2020biobert}, a variant of BERT fine-tuned on biomedical corpora. Finally, for the ECHO encoder, we employ pre-trained ECHO-CLIP~\cite{christensen2024visionechoclip}.

\noindent\textbf{Generalized
Pooling Operator.} To generate probabilistic embeddings, we added two additional layers in parallel on top of each encoder, one for predicting the mean $\mu$ (initialized from the pre-trained backbone) and one for predicting $\log\sigma^2$ (randomly initialized). For computational efficiency, the $\log\sigma^2$ head is a single-layer network, and features are aggregated using Generalized Pooling Operator (GPO) following~\cite{chun2023improvedpcmepp, chen2021learningGPO}.

\noindent\textbf{Batch Normalization.} 
Before GPO for feature aggregation, we incorporate Batch Normalization (BN).  BN enforces a zero mean and unit variance across mini-batch, which may stabilize training and mitigate internal covariate shifts~\cite{ioffe2015batch}. This may be beneficial in probabilistic settings, where BN could ensure estimated distribution parameters remain \textbf{less sensitive} to fluctuations in individual samples.  

\subsection{Binding Multimodal Probabilistic Embeddings}
\label{sec:bindingmodal}

\textsc{ProbMED} integrates CXR--TEXT, ECG--TEXT, ECHO--TEXT, and CXR--ECG pairs in the MIMIC datasets to learn a unified joint embedding space, the available modality pairs is defined as $B\in \{\mathrm{(CXR,TEXT)}, (\mathrm{ECG,TEXT)}$, $\mathrm{(ECHO,TEXT)}, \mathrm{(CXR,ECG)}\}$. CXR--ECG was the only non-text modality pair we trained on as our training datasets had $<1000$ samples of other non-text modality pairs. 
The model is trained using a meta-learning strategy akin to~\cite{girdhar2023imagebind}, where each gradient update is derived from a distinct objective for each available pair (described in $\S$\ref{subsec:finalloss}).

We encode an input for each modality $m$ into a probabilistic embedding using Eq.~\ref{eq:defined_distribution}, as $\mu_{m},\ \sigma_{m} \in \mathbb{R}^D$, and 
\begin{equation}
\label{eq:prob_embed}
    Z_{m} \sim \mathcal{N}\Bigl(\mu_{m},\, \operatorname{diag}(\sigma_{m}^2)\Bigr),
\end{equation}
where $m\in\{\mathrm{CXR}, \mathrm{ECG},\mathrm{ECHO},\mathrm{TEXT}\}$.
\\

\noindent\textbf{Modality-Text Alignment.} For modality-text alignment (e.g., CXR--TEXT), we use InfoNCE loss~\cite{oord2018representation} on the probabilistic embeddings. Let $q_{n}$ be the output of the \textbf{non-text} modality $n\in\{\mathrm{CXR}, \mathrm{ECG},\mathrm{ECHO}\}$, and $k_{t}$ be the output of the \textbf{corresponding paired} text. Based on~\cref{eq:prob_embed}:
\begin{equation} 
\label{eq:cxr_embedding}
    q_{n} \sim \mathcal{N}\Bigl(\mu_{n},\, \operatorname{diag}({\sigma}_{n}^2)\Bigr) \text{ and } k_{t} \sim \mathcal{N}\Bigl(\mu_{t},\, \operatorname{diag}({\sigma}_{t}^2)\Bigr)
\end{equation}

where $t$ refers to the text representation of modality $n$. Then, given a batch of $N$ modality–text pairs $\{(q_{n,i}, k_{t,i})\}_{i=1}^N $, we can calculate the InfoNCE loss as:
\begin{equation}
\label{eq:infoNCE_modality}
    \mathcal{L}_{\mathrm{MOD}_{n,t}} = -\frac{1}{N}\sum_{i=1}^{N} \log \frac{\exp\bigl(-\mathrm{PS}(q_{n,i}, k_{t,i})/\tau\bigr)}{\sum_{j=1}^{N} \exp\bigl(-\mathrm{PS}(q_{n,i}, k_{t,j})/\tau\bigr)},
\end{equation}
where $\tau$ is the temperature and $\mathrm{PS}(\cdot,\cdot)$ is a similarity function computing the similarity between two \textit{probability distributions}. We use symmetric loss: $\mathcal{L}_{\mathrm{MOD}_{n,t}}+\mathcal{L}_{\mathrm{MOD}_{t,n}}$.

\noindent\textbf{Non-text Modality Alignment.} Where possible (\cref{fig:framework}), we enforced consistency across \textit{non-text} modality pairs alignment, termed \textbf{non-text modality}. $\mathcal{L}_{{\mathrm{MOD}_{n,n'}}}$ represents this alignment for $n\neq n'$ using the same equation as~\cref{eq:infoNCE_modality} by replacing $t$ to $n'$. We trained \textsc{ProbMED} on ECG-CXR non-text modality (see $\S$\ref{sec:bindingmodal}). 

\noindent\textbf{Probabilistic Similarity Function.} We adopt $1-H$, where $H$ is the Hellinger distance~\cite{pardo2018statistical}, to measure the similarity between probabilistic embeddings because it is symmetric and bounded, making it well-suited for contrastive learning. The squared Hellinger distance between two multivariate Gaussian distributions~\cite{pardo2018statistical}, $q_n \text{ and } k_t$, follows from \cref{eq:premise}, where the covariance matrices are \(\operatorname{diag}(\sigma_n^2) = \Sigma_n\) and \(\operatorname{diag}(\sigma_t^2) = \Sigma_t\). Then:
\begin{equation} 
\label{eq:closedform_hellinger}
\begin{split}
& H^2(q_n,k_t)= 1 - \frac{\det(\boldsymbol{\Sigma}_n)^{\frac{1}{4}}\det(\boldsymbol{\Sigma}_t)^{\frac{1}{4}}}{
\det\!\Bigl(\frac{\boldsymbol{\Sigma}_n+\boldsymbol{\Sigma}_t}{2}\Bigr)^{\frac{1}{2}}}
\\&
\quad\times \exp\Bigr((-\frac{1}{8}(\mu_n-\mu_t)^\top 
\Bigl(\frac{\boldsymbol{\Sigma}_n+\boldsymbol{\Sigma}_t}{2}\Bigr)^{-1} (\mu_n-\mu_t)\Bigr).
\end{split}
\end{equation}
\noindent We can then simplify the squared Hellinger distance as a product over the $D$ dimensions:
\begin{equation} 
\label{eq:our_hellinger}
\begin{split}
&H^2(q_n,k_t) = 1 - \\& \prod_{o=1}^{D} \left[ \left(\frac{{2\sigma_{n,o}\sigma_{t,o}}}{{{\sigma_{n,o}^2+\sigma_{t,o}^2}}}\right)^{\frac{1}{2}} \exp\!\left(-\frac{(\mu_{n,o}-\mu_{t,o})^2}{4(\sigma_{n,o}^2+\sigma_{t,o}^2)}\right) \right].
\end{split}
\end{equation}
\\
\noindent Following this, we define the similarity measure as $\mathrm{PS}(q_n, k_t) = 1 - \sqrt{H^2(q_n, k_t)}$, converting the squared Hellinger distance into a similarity metric. The details of calculating and computing Hellinger distance are described in the Supplementary $\S$\ref{subsec:hellingercalc}. Unlike alternatives such as the PCME++ closed-form distance (CSD) and the Bhattacharyya distance, the Hellinger distance is symmetric and \textbf{bounded}, stabilizing training by mitigating excessive gradient magnitudes, an essential feature when handling noisy and uncertain nature of medical data. Furthermore, its sensitivity to differences in the means and variances of distributions enables it to capture subtle discrepancies between modalities. Empirically, our findings indicate that the Hellinger similarity facilitates superior convergence and creates a more discriminative joint embedding space.

\subsection{Within-Modality Probabilistic Embeddings}
\label{sec:intramodal}
To make the probability distributions of each modality robust we further propose \textbf{Synthetic Instance Sampling} \textbf{(SIS) Loss}. SIS loss encourages learning meaningful distributions by maximizing the similarity between two sampled instances from the same latent distributions. We adopt the reparameterization trick for multivariate Gaussian with a diagonal covariance structure from~\cite{kingma2013autoVAE}. So, for an input with latent distribution (\cref{eq:defined_distribution}) a sample is obtained:
\begin{equation} \label{eq:reparam_sampling}
    \resizebox{\columnwidth}{!}{$
    z_m^{l} = \mu_m + \operatorname{diag}({\sigma_m}) {\epsilon}^{l}, \quad {\epsilon}^{l} \sim \mathcal{N}\bigl({0},\,\mathbf{I}\bigr),\quad l = \{1,\ldots, N_s\},
    $} %
\end{equation}
where $N_s$ is the number of sampled instances, $\mathbf{I}$ is a ${D \times D}$ identity matrix, and $\epsilon^{l}\in\mathbb{R}^D$. Note: $\sigma_m$ is the modality, $m$, standard deviation. We reformulate InfoNCE~\cite{oord2018representation} to operate on these sampled embedding instances. For each instance $ z_m^l $ and $ z_m^{l'}$, where $l\neq l'$. 
Negative keys are drawn from other instances in the mini-batch. Thus, \textbf{SIS loss}:
\begin{equation}
\label{eq:infoNCE_sampling}
    \mathcal{L}_{\mathrm{SIS}_m} = -\sum_{i=1}^{N} \log \frac{\exp\bigl(-\mathrm{CS}(z^l_{m,i}, z^{l'}_{m,i})/\tau\bigr)}{\sum_{j=1,j\neq i}^{2N} \exp\bigl(-\mathrm{CS}(z^l_{m,i}, z^{l}_{m,j})/\tau\bigr)},
\end{equation}
\noindent where $\mathrm{CS}(\cdot,\cdot)$ denotes the cosine similarity function. We present $\mathcal{L}_\mathrm{SIS}$ for $N_s=2$, which is analogous to SimCLR~\cite{chen2020simplesimclr}. Thus, these two samples act as distinct instantiated \textit{views} of the same distribution, reinforcing the distributions to maintain the meaningful variance. $\mathcal{L}_\mathrm{SIS}$ encourages the model to learn distributions by enforcing consistency between samples drawn from the same underlying distribution. We used $\mathcal{L}_\mathrm{SIS}$ on \textbf{all} modalities during pretraining. 

\begin{table*}[t]
    \centering
    \label{tab:retrieval_performance}    
    \begin{subtable}[t]{\textwidth} 
        \centering
        \caption{\textbf{TEXT-to-CXR retrieval.}}
        \label{tab:cxr_recall}
        \resizebox{0.70\textwidth}{!}{ 
        \begin{tabular}{l?cc?cc?cc?cc?c}
            \toprule
            \multirow{2}{*}{\textcolor{Green}{ }} & \multirow{2}{*}{Prob?} & \multirow{2}{*}{Similarity} 
            &\multicolumn{2}{c?}{\textcolor{Green}{MIMIC-CXR}} 
            & \multicolumn{2}{c?}{\textcolor{Green}{OpenI}}
            & \multicolumn{2}{c?}{\textcolor{Green}{Chexpert5x200}} 
            & \multirow{2}{*}{\textcolor{Green}{RSUM}} \\ 
            & & & R@1 & R@5 & R@1 & R@5 & R@1 & R@5 \\  
            \hline
            MedCLIP~\cite{wang2022medclip} & \textcolor{Gray}{\xmark}& Cosine   & 1.0 & 4.3 & 0.6 & 2.8 & 2.6 & 3.0 & 14.3\\
            CXR-CLIP~\cite{you2023cxrclip} & \textcolor{Gray}{\xmark} & Cosine & \underline{47.3} & \underline{70.4} & \textbf{12.7} & 25.2 & 8.5 & 23.0 & \underline{187.1} \\
            BiomedCLIP~\cite{zhang2023biomedclip} & \textcolor{Gray}{\xmark} & Cosine &  36.2 & 59.9 & 9.0 & 19.9 & 6.4 & 19.8 & 151.2 \\
            CheXzero~\cite{tiu2022expertchexzero} & \textcolor{Gray}{\xmark}& Cosine  & 26.7 & 50.0 & 5.8 & 15.1 & 3.5 & 17.8 & 118.9\\
            MEDBind~\cite{gao2024medbind} & \textcolor{Gray}{\xmark}& Cosine  & 40.8 & 67.5 & \underline{11.6} & 25.5 & 7.9 & 21.4 & 174.7 \\
            
            BioVil-T~\cite{bannur2023learningbiovil} &  \textcolor{Gray}{\xmark}& Cosine & 28.4 & 58.2 & 8.1 & 18.9 & 4.9 & 17.1 & 135.6\\
            SAT~\cite{liu2023improvingSAT} &  \textcolor{Gray}{\xmark}& Cosine & 40.3 & 69.2 & 6.7 & 14.7 & \textbf{9.1} & \underline{26.7} & 166.7 \\
            \hline
            PCME++~\cite{chun2023improvedpcmepp} & \cmark & CSD  & 32.1 & 50.6 & 10.8 & \textbf{28.3} & 4.0 & 16.2 & 142.0 \\
            \textsc{ProbMED} (Ours) & \cmark & Hellinger & \textbf{47.9}& \textbf{71.4} & \textbf{12.7} & \underline{27.5} & \underline{8.9} & \textbf{28.4} & \textbf{196.8}\\
            \bottomrule
        \end{tabular}}
    \end{subtable}

    \vspace{4pt} 

    \begin{subtable}[t]{0.51\textwidth} 
        \centering
        \caption{\textbf{TEXT-to-ECG retrieval.}}
        \label{tab:ecg_recall}
        \resizebox{\textwidth}{!}{ 
        \begin{tabular}{l?c?cc?cc?c}
            \toprule
            \multirow{2}{*}{\textcolor{Purple}{ }} &\multirow{2}{*}{Similarity} 
            & \multicolumn{2}{c?}{\textcolor{Purple}{MIMIC-ECG}} 
            & \multicolumn{2}{c?}{\textcolor{Purple}{PTB-XL}} 
            & \multirow{2}{*}{\textcolor{Purple}{RSUM}} \\ 
            & & R@1 & R@5 & R@1 & R@5 \\  
            \hline
            ECG-CLIP~\cite{gao2024medbind} & Cosine & 40.8 & 76.7 & 2.3 & 9.8 & 129.6\\
            MEDBind~\cite{gao2024medbind} & Cosine & \underline{44.1} & \underline{78.2} & \textbf{3.1} & \underline{12.1} & \underline{137.5}\\
            \hline
            PCME++~\cite{chun2023improvedpcmepp} & CSD & 40.9 & 54.8 & 1.3 & 11.3 & 108.3\\
            \textsc{ProbMED} (Ours) & Hellinger & \textbf{48.3} & \textbf{87.0} & \underline{2.8} & \textbf{12.2} & \textbf{150.3}\\
            \bottomrule
        \end{tabular}}
    \end{subtable}%
    \hspace{5pt} 
    \begin{subtable}[t]{0.40\textwidth} 
        \centering
        \caption{\textbf{TEXT-to-ECHO retrieval.}}
        \label{tab:echo_recall}
        \resizebox{\textwidth}{!}{ 
        \begin{tabular}{l?c?cc?c}
            \toprule
            \multirow{2}{*}{\textcolor{Purple}{ }} & \multirow{2}{*}{Similarity} 
            & \multicolumn{2}{c?}{\textcolor{Maroon}{MIMIC-ECHO}}
            & \multirow{2}{*}{\textcolor{Maroon}{RSUM}} \\ 
            & & R@1 & R@5 \\  
            \hline
            EchoCLIP~\cite{christensen2024visionechoclip}  & Cosine & \underline{1.1} & \underline{6.4} & \underline{7.5} \\
            \hline
            PCME++~\cite{chun2023improvedpcmepp} & CSD & 1.0 & 5.0 & 6.0 \\
            \textsc{ProbMED} (Ours)  & Hellinger & \textbf{2.4} & \textbf{7.8} & \textbf{10.2} \\
            \bottomrule
        \end{tabular}}
    \end{subtable}
    \caption{Cross-modal retrieval performance, Recall@K, for (a) TEXT-to-CXR, (b) TEXT-to-ECG, and (c) TEXT-to-ECHO retrieval tasks. CSD stands for closed-form distance from~\cite{chun2023improvedpcmepp}. The best performance is in \textbf{bold}, while the second-best is \underline{underlined}.}
\end{table*}
\begin{table}
  \centering
  \resizebox{\columnwidth}{!}{
  \begin{tabular}{l|cccc}
    \thickhline
Dataset & Emerg. & Task & \#Cls & \#test \\ 
\hline
\textcolor{Green}{MIMIC-CXR}~\cite{johnson2019mimiccxr}$^*$ & \textcolor{Gray}{\xmark} & Retrieval/Multimodal & - & 24,799 \\
\textcolor{Green}{OpenI}~\cite{demner2016preparing_openi} & \textcolor{Gray}{\xmark} & Retrieval & - & 2,864 \\
\textcolor{Green}{CheXpert5x200}~\cite{irvin2019chexpert} & \textcolor{Gray}{\xmark} & Retrieval & 5 & 1,000 \\
\textcolor{Green}{RSNA}~\cite{shih2019augmenting_rsna} & \textcolor{Gray}{\xmark} & Classification & 2 & 5,338 \\
\textcolor{Green}{COVID Kaggle}~\cite{chowdhury2020can_kaggle_covid1} & \textcolor{Gray}{\xmark} & Classification & 2 & 2,780 \\
\textcolor{Green}{Montgomery}~\cite{candemir2013lungmontgomery} & \textcolor{Gray}{\xmark} & Classification & 2 & 106  \\
\rowcolor{CornflowerBlue!10} 
\textcolor{Green}{CheXchoNet}~\cite{bhave2024deepchexcho} & \cmark & Classification & 2 & 3,667 \\

\hline
\textcolor{Purple}{MIMIC-ECG}~\cite{gow_mimicecg}$^*$ & \textcolor{Gray}{\xmark} & Retrieval/Multimodal & - & 24,644 \\
\textcolor{Purple}{PTB-XL}~\cite{strodthoff2020deep_ptbxl} & \textcolor{Gray}{\xmark} & Retrieval/Classification & 71 & 2,198 \\
\textcolor{Purple}{ICBEB}~\cite{liu2018open_icbeb} & \textcolor{Gray}{\xmark} & Classification & 9 & 1,376 \\
\rowcolor{CornflowerBlue!10} 
\textcolor{Purple}{MUSIC}~\cite{martinmusic} & \cmark & Classification & 2 & 125 \\
\hline

\textcolor{Maroon}{MIMIC-ECHO}~\cite{gow2023mimicecho}$^*$ & \textcolor{Gray}{\xmark} & Retrieval & - & 1,957 \\
\textcolor{Maroon}{EchoNet-Dynamic}~\cite{ouyang2019echonet} & \textcolor{Gray}{\xmark} & Classification & 2 & 1,264  \\
    \thickhline
  \end{tabular}}
  \caption{Datasets for \textcolor{Green}{CXR}, \textcolor{Purple}{ECG}, and \textcolor{Maroon}{ECHO} modalities. For each dataset, we report the task (i.e., classification, retrieval, and multimodal), number of classes (\#Cls), and number of test samples (\#test). \colorbox{CornflowerBlue!10}{CheXchoNet} and \colorbox{CornflowerBlue!10}{MUSIC} are both \textbf{emergent} (Emerg.) and external datasets. \textsc{ProbMED} was never trained on CXR-ECHO or ECG-ECHO pairs. $^*$These datasets were used for pertaining, but test set was reserved (Supplementary $\S$\ref{suppsec:datasets})}
  \label{tab:external_data}
\end{table}

\subsection{Variational Information Bottleneck}
To prevent the collapse of the variance, we use a Variational Information Bottleneck (VIB) loss~\cite{ohmodelingVIB} similar to~\cite{chun2023improvedpcmepp}. VIB loss is computed as KL-divergence between learned latent distribution $Z_m$, \cref{eq:prob_embed} and a standard normal prior:
\begin{equation} \label{eq:VIB_loss}
    \mathcal{L}_{\mathrm{VIB}_m} = -\frac{1}{N}\sum_{i=1}^{N}{\mathrm{KL} \!\Bigl(Z_{m,i} \,\Big\|\, \mathcal{N}(\mathbf{0},\, \mathbf{I})\Bigr)}.
\end{equation}
Empirically, VIB loss prevents variance collapse.

\subsection{Final Loss Function}
\label{subsec:finalloss}
The overall loss is a weighted sum of multiple components. We randomly sample a modality pair from the available pairs for each gradient update. $\mathcal{L}_\mathrm{SIS}$ and  $\mathcal{L}_{\mathrm{VIB}}$ are always included. So for $\{(m1,m2)\}\in B$, where $B$ is the set of available modality pairs (defined in $\S$\ref{sec:bindingmodal}), then:
\begin{equation} \label{eq:final_loss}
    \begin{split}
    &\mathcal{L}_{{m1,m2}} = \alpha(\mathcal{L}_{\mathrm{MOD}_{m1,m2}} + \mathcal{L}_{\mathrm{MOD}_{m2,m1}}) \\&+\beta( \mathcal{L}_{\mathrm{SIS}_{m1}}+\mathcal{L}_{\mathrm{SIS}_{m2}}) + \gamma \ (\mathcal{L}_{{\mathrm{VIB}_{m1}}}+\mathcal{L}_{{\mathrm{VIB}_{m2}}}),
    \end{split}
\end{equation}
The weights of each loss and $\tau$ scaling were set empirically and are presented and are presented in Supplementary $\S$\ref{suppsec:pretraining}.   
\section{Experiments and Results}
This section introduces our experiments to evaluate \textsc{ProbMED}, compared to other Med-VLPMs. We trained \textsc{ProbMED} exclusively on MIMIC datasets. We also trained a PCME++ model for comparison. We evaluated on the 3 MIMIC and 10 external datasets. All datasets used in testing are in~\cref{tab:external_data}, and detailed in Supplementary $\S$\ref{suppsec:datasets}.

\noindent\textbf{Text-to-Modality Retrieval:} 
In text-based retrieval experiments on free-form clinical notes, deterministic embeddings often miss linguistic nuances. Our probabilistic framework represents each text as a distribution, enabling similarity metrics that account for central tendency and uncertainty through probabilistic similarity comparisons.

\noindent\textbf{Zero-shot and few-shot classification:}  We tested \textsc{ProbMED} on traditional zero-shot (ZS) and few-shot (FS) datasets. In addition, we explored emergent ZS and FS classification, which refers to a model's ability to align modality pairs that were not explicitly trained during pretraining (e.g., ECG and ECHO pairs were unseen). Inspired by~\cite{girdhar2023imagebind}, we observed that training this unified model, \textsc{ProbMED}, facilitated these emergent alignments. We illustrated this phenomenon by evaluating classification tasks that map ECG or CXR inputs to \textit{ECHO labels}, even though these specific pairs were never observed together during training.\par
\begin{table*}[t]
\centering
\label{tab:allzero_fewshot} 
\begin{subtable}[t]{\textwidth} 
\centering
\caption{\textbf{CXR-based zero and few-shot}}
\label{tab:cxrfewshot}
\resizebox{0.8\textwidth}{!}{ 
\begin{tabular}{l?ccc?ccc?ccc?ccc|cc}
 \thickhline
\multirow{2}{*}{} & \multicolumn{3}{c?}{\textcolor{Green}{COVID Kaggle}} & \multicolumn{3}{c?}{\textcolor{Green}{RSNA}} & \multicolumn{3}{c?}{\textcolor{Green}{Montgomery}} & \multicolumn{3}{c|}{\textcolor{Green}{\cellcolor{CornflowerBlue!7}CheXchoNet $\star$}} &\multicolumn{2}{c}{\textcolor{Green}{Overall Rank}}\\ 
& ZS & 4S& 16S& ZS & 4S& 16S& ZS & 4S& 16S& \cellcolor{CornflowerBlue!7}ZS & \cellcolor{CornflowerBlue!7}4S& \cellcolor{CornflowerBlue!7}16S & ZS & FS \\
\hline
MedCLIP~\cite{wang2022medclip} & 75.3& 85.5& 90.8& 75.7& 58.0& 65.4& \underline{88.3}& 87.3& 88.5& \cellcolor{CornflowerBlue!7}61.6& \cellcolor{CornflowerBlue!7}55.8& \cellcolor{CornflowerBlue!7}63.9 & 4 & 8\\
CXR-CLIP~\cite{you2023cxrclip} & 76.9& \textbf{86.7} & 91.6& 72.7& 64.1& 70.9& 81.1& 85.8& 91.6& \cellcolor{CornflowerBlue!7}61.1& \cellcolor{CornflowerBlue!7}53.2& \cellcolor{CornflowerBlue!7}59.7 & 6 & 7\\
BiomedCLIP~\cite{zhang2023biomedclip} & \underline{84.4}& 86.0& 89.4& 81.6& \underline{80.3} & 84.0 & 84.5& 87.0 & 92.2& \cellcolor{CornflowerBlue!7}62.1 & \cellcolor{CornflowerBlue!7}59.8 & \cellcolor{CornflowerBlue!7}61.8 & 3 & 2\\
CheXzero~\cite{tiu2022expertchexzero}&  79.5 & 82.8 & 88.4&  47.9& 75.0 & 82.7& 71.6& 88.5& \underline{92.9} & \cellcolor{CornflowerBlue!7}\underline{67.9}& \cellcolor{CornflowerBlue!7}\underline{60.3} & \cellcolor{CornflowerBlue!7}\underline{66.1} & 8 & 3\\
MEDBind~\cite{gao2024medbind} & \textbf{86.4}& 86.2 & \textbf{92.0}& 80.0& 67.3& 73.4& 86.8 & \underline{89.9}& 91.8& \cellcolor{CornflowerBlue!7}62.0& \cellcolor{CornflowerBlue!7}57.6& \cellcolor{CornflowerBlue!7}65.4 & 2 & 4\\
BioViL-T~\cite{bannur2023learningbiovil} & 69.9 & 67.7 & 76.6 & 71.5 & \textbf{82.2} & \textbf{85.9} & \textbf{88.6} & 88.6 & 91.8 & \cellcolor{CornflowerBlue!7}63.3 & \cellcolor{CornflowerBlue!7}56.6 & \cellcolor{CornflowerBlue!7}64.7 & 5 & 5\\
SAT~\cite{liu2023improvingSAT} & 72.1 & 81.8 & 87.4 & 73.5 & 56.0 & 61.2 & 75.9 & 78.9 & 82.3 & \cellcolor{CornflowerBlue!7}60.3 & \cellcolor{CornflowerBlue!7}56.0 & \cellcolor{CornflowerBlue!7}65.9 & 7 & 10 \\
MedKLIP$^\dagger$~\cite{wu2023medklip} & 68.8 & 85.4 & 90.2 & \underline{82.4} & 79.3 & 82.8 & 51.7 & 79.5 & 84.2 & \cellcolor{CornflowerBlue!7}63.8 & \cellcolor{CornflowerBlue!7}56.0 & \cellcolor{CornflowerBlue!7}63.6 & 9 & 5\\
\hline
PCME++~\cite{chun2023improvedpcmepp}& 48.6& 79.6& 85.9& 45.2& 73.2& 79.2& 50.4& 75.7& 81.8& \cellcolor{CornflowerBlue!7}59.9& \cellcolor{CornflowerBlue!7}56.8& \cellcolor{CornflowerBlue!7}62.7 & 10 & 9\\
\textsc{ProbMED} (Ours)& \textbf{86.4}& \underline{86.5} & \underline{91.8}& \textbf{82.5} & \textbf{82.2}& \underline{84.7}& 84.3& \textbf{93.1} & \textbf{93.7}& \cellcolor{CornflowerBlue!7}\textbf{69.5} & \cellcolor{CornflowerBlue!7}\textbf{63.3} & \cellcolor{CornflowerBlue!7}\textbf{68.5} & 1 & 1\\
 \thickhline
\end{tabular}}
\end{subtable}

\vspace{4pt} 
\begin{subtable}[t]{0.65\textwidth} 
\centering
\caption{\textbf{ECG-based zero and few-shot}}
\label{tab:ecgfewshot}
\resizebox{\textwidth}{!}{ 
\begin{tabular}{l?ccc?ccc?ccc|cc}
 \thickhline
\multirow{2}{*}{} & \multicolumn{3}{c?}{\textcolor{Purple}{PTB-XL}} & \multicolumn{3}{c?}{\textcolor{Purple}{ICBEB}} & \multicolumn{3}{c|}{\cellcolor{CornflowerBlue!7}\textcolor{Purple}{MUSIC $\star$}} &\multicolumn{2}{c}{\textcolor{Purple}{Overall Rank}}\\ 
& ZS & 4S & 16S& ZS & 4S & 16S& \cellcolor{CornflowerBlue!7}ZS & \cellcolor{CornflowerBlue!7}4S & \cellcolor{CornflowerBlue!7}16S & ZS & FS\\
\hline
ECG-CLIP~\cite{gao2024medbind} & 52.3& 67.1& 71.2& 61.9& 69.1& 74.1& \cellcolor{CornflowerBlue!7}60.4& \cellcolor{CornflowerBlue!7}48.6& \cellcolor{CornflowerBlue!7}51.4 & 4 & 5\\
MEDBind~\cite{gao2024medbind}& 55.3& 71.1& \underline{81.8}& 65.7& 81.2 & \underline{87.8}& \cellcolor{CornflowerBlue!7}61.3& \cellcolor{CornflowerBlue!7}\underline{51.5}& \cellcolor{CornflowerBlue!7}\underline{54.6} &3 & 2\\
ECG-FM~\cite{mckeen2024ecg}& -& 69.1& 71.6& -& 69.3& 71.8& \cellcolor{CornflowerBlue!7}-& \cellcolor{CornflowerBlue!7}50.0& \cellcolor{CornflowerBlue!7}53.1 & - & 4\\
\hline
PCME++~\cite{chun2023improvedpcmepp}& \underline{61.2}& \underline{75.4}& 79.9& \underline{71.3}& \underline{74.1}& 80.5& \cellcolor{CornflowerBlue!7}\underline{67.2}& \cellcolor{CornflowerBlue!7}46.7& \cellcolor{CornflowerBlue!7}48.6 & 2 & 3\\
\textsc{ProbMED} (Ours) & \textbf{64.5} & \textbf{82.6} & \textbf{87.6} & \textbf{75.3} & \textbf{84.8} & \textbf{90.1} & \cellcolor{CornflowerBlue!7}\textbf{70.5} & \cellcolor{CornflowerBlue!7}\textbf{53.8} & \cellcolor{CornflowerBlue!7}\textbf{59.1} &1 &1 \\
 \thickhline
\end{tabular}

}

\end{subtable}
\hspace{5pt} 
\begin{subtable}[t]{0.33\textwidth} 
\centering
\caption{\textbf{ECHO-based zero and few-shot}}
\label{tab:echofewshot}
\resizebox{\textwidth}{!}{ 
\begin{tabular}{l?ccc|cc}
 \thickhline
\multirow{2}{*}{} &\multicolumn{3}{c|}{\textcolor{Maroon}{EchoNet-Dynamic}}&\multicolumn{2}{c}{\textcolor{Maroon}{Overall Rank}}\\ 
 & ZS & 4S & 16S & ZS & FS\\
\hline
EchoCLIP~\cite{christensen2024visionechoclip}& \underline{75.1} & \textbf{88.3} & \underline{95.0} & 2 & 2\\
\hline
PCME++~\cite{chun2023improvedpcmepp} & 73.6 & 87.5 & 94.1 & 3 & 3\\
\textsc{ProbMED} (Ours) & \textbf{82.6} & \underline{87.7} & \textbf{96.2} & 1 &1 \\
 \thickhline
\end{tabular}}
\end{subtable}
\caption{Performance comparison of \textsc{ProbMED} with state-of-the-art Med-VLPMs on many different modalities, few-shot tasks across (a) CXR-based, (b) ECG-based, and (c) ECHO-based datasets. We report AUROC scores under zero-shot (ZS), 4-shot (4S), and 16-shot (16S). For probabilistic models (i.e., PCME++ and \textsc{ProbMED}), only $\mu$ embeddings were used in few-shot learning for fair comparisons. {\textbf{\textcolor{Green}{$\star$}}\textbf{CheXchoNet}} is an \textit{emergent} dataset using CXR as an input to predict an ECHO label, composite of severe left ventricular hypertrophy (SLVH) and dilated left ventricle (DLV). {\textbf{\textcolor{Purple}{$\star$}}\textbf{MUSIC}} is an \textit{emergent} dataset using ECG (input) to ECHO label, composite of SLVH and DLV. $^\dagger$MedKLIP performance was with the 77 classes in disease book (i.e., COVID and CheXchoNet were added to the original disease book).}
\end{table*}

\noindent\textbf{Multimodal Classification:} Medical decision-making relies on multiple sources of evidence (e.g., combining information from both CXR and ECG for more accurate prognoses~\cite{fonseca2004valuemultimodal}). As discussed in the introduction, a central motivation behind \textsc{ProbMED} is its capacity to encode and integrate information from various modalities. We demonstrated this by evaluating the ZS and FS performance of \textsc{ProbMED} using embeddings from multiple modalities and comparing its performance to that of traditional approaches with a single modality.\par
\noindent\textbf{Pushing the Boundaries of Probabilistic Models:}
While prior approaches~\cite{you2023cxrclip,zhang2023biomedclip} emphasize improved ZS/FS performance using deterministic embeddings, \textsc{ProbMED} leverages probabilistic embeddings to capture each modality's intrinsic uncertainty and distributional characteristics. In addition to the conventional way of doing ZS and FS, we underlined use-cases that leverage the \textit{learned} \textit{distributions} to improve ZS and FS performance. \textbf{1) ZS}: We used an uncertainty-based prompt filtering mechanism proposed in~\cite{chun2023improvedpcmepp}, which filters out prompts with high uncertainty, assuming more ambiguous prompts are therefore less helpful as classifier references. \textbf{2) FS}: the probabilistic nature of \textsc{ProbMED} is leveraged to generate synthetic samples that provide more training samples. For example, by sampling from the latent distribution, a \textit{single} image can generate \textit{multiple} effective training examples for FS training. \par

\subsection{Text-to-Modality Retrieval}
We benchmarked \textsc{ProbMED} across several text-to-modality retrieval tasks against state-of-the-art Med-VLPMs. Our evaluation leveraged a probabilistic framework that measures central tendency and uncertainty using tailored similarity metrics. For all models, we used the most appropriate similarity for the respective model (i.e., cosine similarity was used for deterministic methods, while PCME++ and \textsc{ProbMED} used probabilistic similarities).
\cref{tab:cxr_recall},~\cref{tab:ecg_recall}, and~\cref{tab:echo_recall} summarizes the top-k performance (Recall@K) across three retrieval tasks: TEXT-to-CXR, TEXT-to-ECG, and TEXT-to-ECHO. Compared to competing models, \textsc{ProbMED} achieved the highest overall recall performance in these tasks, as indicated by the recall total sum (RSUM). These results underscore the feasibility of incorporating probabilistic modeling and multimodal binding into classical tasks. Using our proposed method and training on multiple modality pairs, we saw that \textsc{ProbMED} improves the modality-text alignment.

\subsection{Zero-shot and few-shot classification}
\label{sec:zerofewshot}
We assessed \textsc{ProbMED} using \textit{traditional} ZS and FS protocols ~\cite{girdhar2023imagebind,radford2021learning_clip} for a fair comparison between probabilistic and deterministic models. ZS performance was calculated using the similarity metrics between text and non-text modality embeddings according to~\cite{girdhar2023imagebind}. Like recall, we applied the most appropriate similarity to measure the relationship between modality-text pairs. We employed linear probing for FS learning following~\cite{girdhar2023imagebind}. Since \textsc{ProbMED} uses probabilistic embeddings, we only used $\mu$ embeddings to enable direct comparison to the output embeddings from the deterministic encoders. See $\S$\ref{sec:probabilistic_use_cases} for leveraging the complete probabilistic distribution.

\cref{tab:cxrfewshot},~\cref{tab:ecgfewshot}, and~\cref{tab:echofewshot} summarizes the area under the receiver operating characteristic curve (AUROC) of ZS and FS results for multiple datasets. \textsc{ProbMED} had competitive results and often outperformed state-of-the-art models across the evaluated modalities. In particular, our method significantly enhances CXR and ECG tasks while marginally surpassing ECHO experiments. In the \textbf{emergent} tasks (CheXchoNet and MUSIC), we noted substantial performance gains with \textsc{ProbMED}, where training on only ECHO-text pairing significantly improves performance.

\subsection{Multimodal Classification}
We evaluated \textsc{ProbMED} for chronic kidney disease (CKD) and chronic heart disease (CHD) classification under ZS and FS settings in MIMIC datasets, particularly on a subset of patients with both CXR and ECG (\textbf{MIMIC-CONNECT}, described in more detail in Supplementary $\S$\ref{subsec:mimicconnect}). In these experiments, we explored using CXR-only, ECG-only, and \textit{concatenation} of CXR and ECG (i.e., multimodal). 

The integration process is visualized in~\cref{fig:example}, showing how CXR and ECG can be used \textit{together} for complex diseases.~\cref{tab:mimic} presents AUROC comparisons among state-of-the-art Med-VLPMs, PCME++, and our \textsc{ProbMED} approach across single-modality (CXR \textit{or} ECG), and multimodality (CXR \textit{and} ECG). Overall, \textsc{ProbMED} outperformed competing methods. CXR+ECG with \textsc{ProbMED} produced gains in tasks, improving single-modality baselines by \textbf{the largest gain of 7.8}\%--in ZS and FS scenarios. These results emphasize the value of leveraging multiple data sources in medical decision-making and highlight \textsc{ProbMED}’s effectiveness in integrating multimodal information for improved prognosis.
\begin{table}[t]

  \centering
  \begin{subtable}[b]{\columnwidth}
      \caption{\textbf{CXR and ECG concatenated to improve prognostication.}}
    \label{fig:example}
    \centering
    \includegraphics[width=0.8\columnwidth]{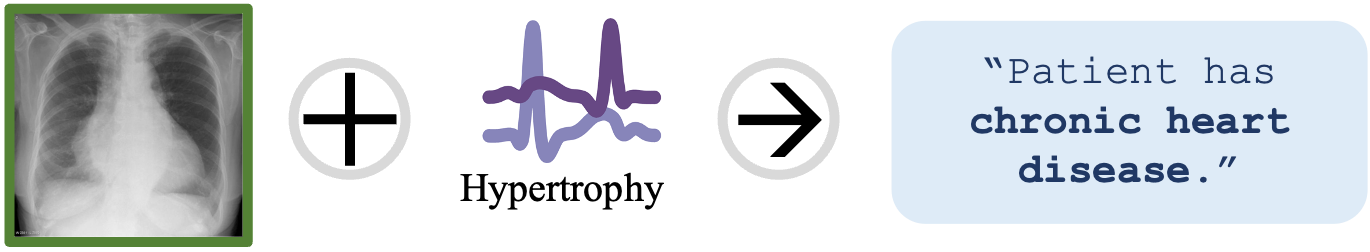}

  \end{subtable}
  
  \vspace{0.5em} 
  
  \begin{subtable}[b]{\columnwidth}
    \caption{\textbf{AUROC performance (in \%) on two prognostic tasks.}}
    \label{tab:mimic}
    \centering
    \resizebox{\columnwidth}{!}{
    \begin{tabular}{l?ccc?ccc?c}
      \thickhline
\multirow{2}{*}{} & \multicolumn{3}{c?}{CKD} & \multicolumn{3}{c?}{CHD} & SUM \\
                  & ZS   & 4S   & 16S  & ZS   & 4S   & 16S  & \\
\hline
\multicolumn{8}{l}{\textit{State-of-the-art best performance from respective VLPMs}} \\
\hline
\rowcolor{gray!7}
CXR         & 73.6\cite{tiu2022expertchexzero} & 66.9\cite{tiu2022expertchexzero} & 71.2\cite{zhang2023biomedclip} & \underline{77.1}\cite{tiu2022expertchexzero} & 68.4\cite{you2023cxrclip} & 75.2\cite{you2023cxrclip} & 432.4 \\
\rowcolor{gray!7}
ECG & 61.2\cite{gao2024medbind} & 66.4\cite{gao2024medbind} & 67.8\cite{gao2024medbind} & 65.7\cite{gao2024medbind}  & 73.5\cite{gao2024medbind} & 74.1\cite{gao2024medbind} & 408.8 \\
\rowcolor{gray!5}
CXR+ECG$^*$  & 71.5\cite{gao2024medbind} & 68.4\cite{gao2024medbind} & 69.8\cite{gao2024medbind} & 75.3\cite{gao2024medbind}  & 71.7\cite{gao2024medbind} & 78.6\cite{gao2024medbind} & 435.3 \\
\hline
\multicolumn{8}{l}{\textit{PCME++}} \\
\hline
CXR         & 51.0 & 68.6 & 68.8 & 51.1  & 72.7 & 76.7 & 388.9 \\
ECG         & 31.7 & 66.7 & 70.3 & 40.3  & \underline{74.7} & 76.7 & 360.4 \\
CXR+ECG     & 46.8 & 69.4 & 71.6 & 52.4  & \textbf{76.9} & 78.7 & 395.8 \\
\hline
\multicolumn{8}{l}{\textsc{ProbMED} \textit{(Ours)}} \\
\hline
CXR         & \underline{75.0} & \underline{70.6} & \underline{76.5} & 77.0  & 71.9 & \underline{79.8} & \underline{450.8} \\
ECG         & 68.5 & 67.4 & 71.1 & 70.3  & 72.2 & 76.7 & 426.2 \\
CXR+ECG     & \textbf{78.1} & \textbf{71.5} & \textbf{76.8} & \textbf{78.4}  & 73.0 & \textbf{80.8} & \textbf{458.6}\\
      \thickhline
    \end{tabular}
    }

  \end{subtable}
  \caption{Multimodal classification results for chronic kidney disease (CKD) and chronic heart disease (CHD). (a) A visual example of multimodal decision-making with two non-text modalities. (b) Classification results under ZS, 4S, and 16S settings. SUM represents the total performance across the ZS and FS settings (in \%). The best Med-VLPM performance is reported in (b) and cited. *Only MEDBind\cite{gao2024medbind} processes both ECG and CXR.}
  \label{fig:stacked}
\end{table} 
\subsection{Pushing the Boundaries of Probabilistic Models}
\label{sec:probabilistic_use_cases}
This section outlines some experimental assessments for the utility of probabilistic modeling. 

\noindent\textbf{Uncertainty-based Prompt Filtering}: We showcased this task as a proof-of-concept for ZS classification in CXR datasets, first proposed in~\cite{chun2023improvedpcmepp}. Although ZS requires a well-curated set of prompts for prediction, the common way is taking the \textit{average} of multiple text prompts to make a robust prototype for each class. However, optimizing the best set of prompts is not trivial. We perform uncertainty-based prompt filtering, where we choose the prompts based on the learned variance, $\sigma^2$, assuming the prompts with low variance are robust. For comparison, we conducted ZS experiments using: (1) a single prompt \texttt{"Chest X-ray of \{$\cdot$\}"} where $\cdot$\ denotes the name of a target disease for each dataset, and (2) 80 prompts describing the disease for the CXR datasets. For the uncertainty-based filtering, we chose $k$ prompts with the lowest variance, which worked the best for each dataset~\cite{chun2023improvedpcmepp}. Our results in~\cref{tab:zerofilter} demonstrate that using prompt filtering on the learned variance consistently improved the ZS performance across all CXR datasets. Thus, \textsc{ProbMED} captures meaningful probability distributions of the given prompts.
\begin{figure}
    \centering
    \includegraphics[width=1\columnwidth]{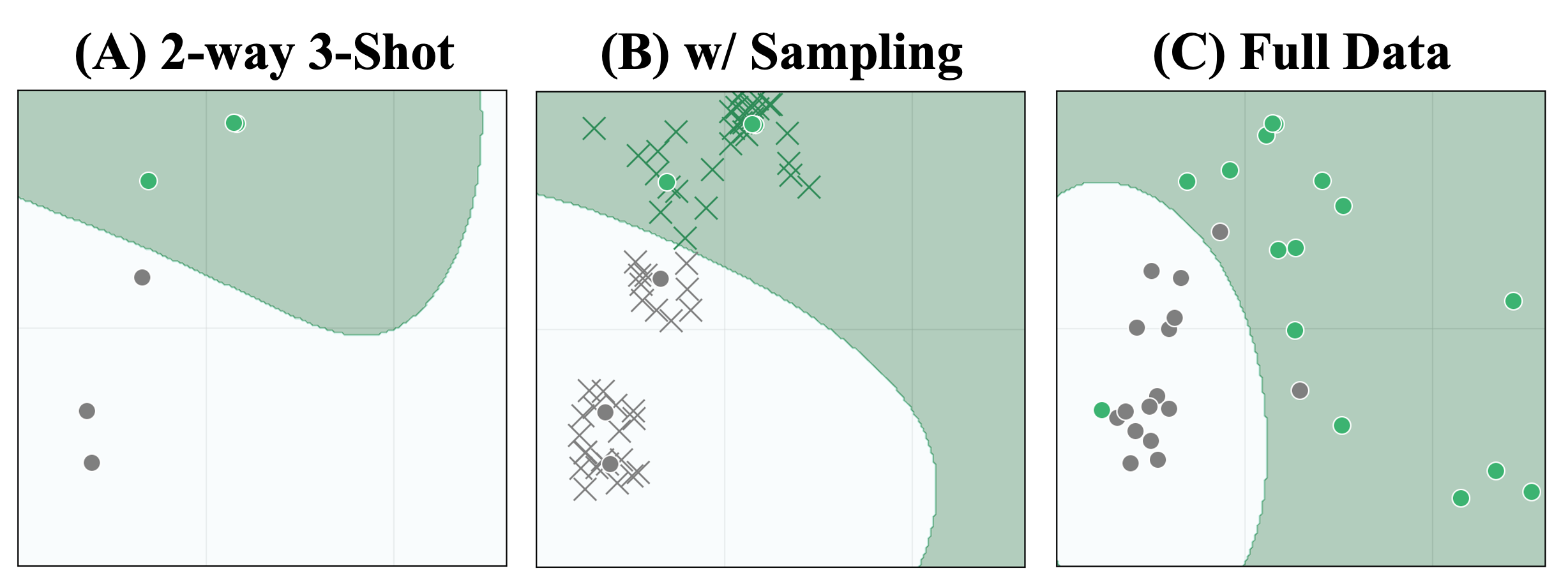}
    \caption{Qualitative comparison of decision boundaries on Kaggle COVID. Each plot shows data projected on the same PCA with an SVM (radial kernel) decision boundary.  (A) The boundary is less reliable in a 2-way 3-shot few-shot setting. (B) \textsc{ProbMED}'s probabilistic sampling (\textbf{X-marks} indicate sampled points) improves the boundary with more samples. (C) Using the full dataset. \textit{Note}: two green points are overlapped in the 3-shot scenario.}
    \label{fig:highlevel_samplingoverview}
\end{figure}
\begin{table}[t]
\centering
\centering
\resizebox{\columnwidth}{!}{%
\begin{tabular}{l?c?c?c?c|c}
\thickhline
Prompts & CX & RN & CV & MG & SUM \\ 
\hline
\texttt{"Chest X-ray of \{$\cdot$\}"} & 70.2 & 63.4 & 78.8 & \underline{86.8} & 299.2 \\
All 80 prompts & \underline{70.8} & \underline{83.0} & \underline{87.2} & 85.6 & \underline{326.6} \\
Uncertainty-based filtering & \textbf{71.0} & \textbf{85.6} & \textbf{87.3} & \textbf{87.7} & \textbf{331.6} \\
\thickhline
\end{tabular}%
}

\caption{Zero-shot performance using uncertainty-based prompt-filtering. \textbf{CX}: CheXchoNet, \textbf{RN}: RSNA, \textbf{CV}: COVID, and \textbf{MG}: Montgomery, \textbf{SUM}: the sum of performance across datasets.}
\label{tab:zerofilter}
\end{table}

\begin{figure*}[t]
    \centering
    \includegraphics[width=0.83\textwidth]{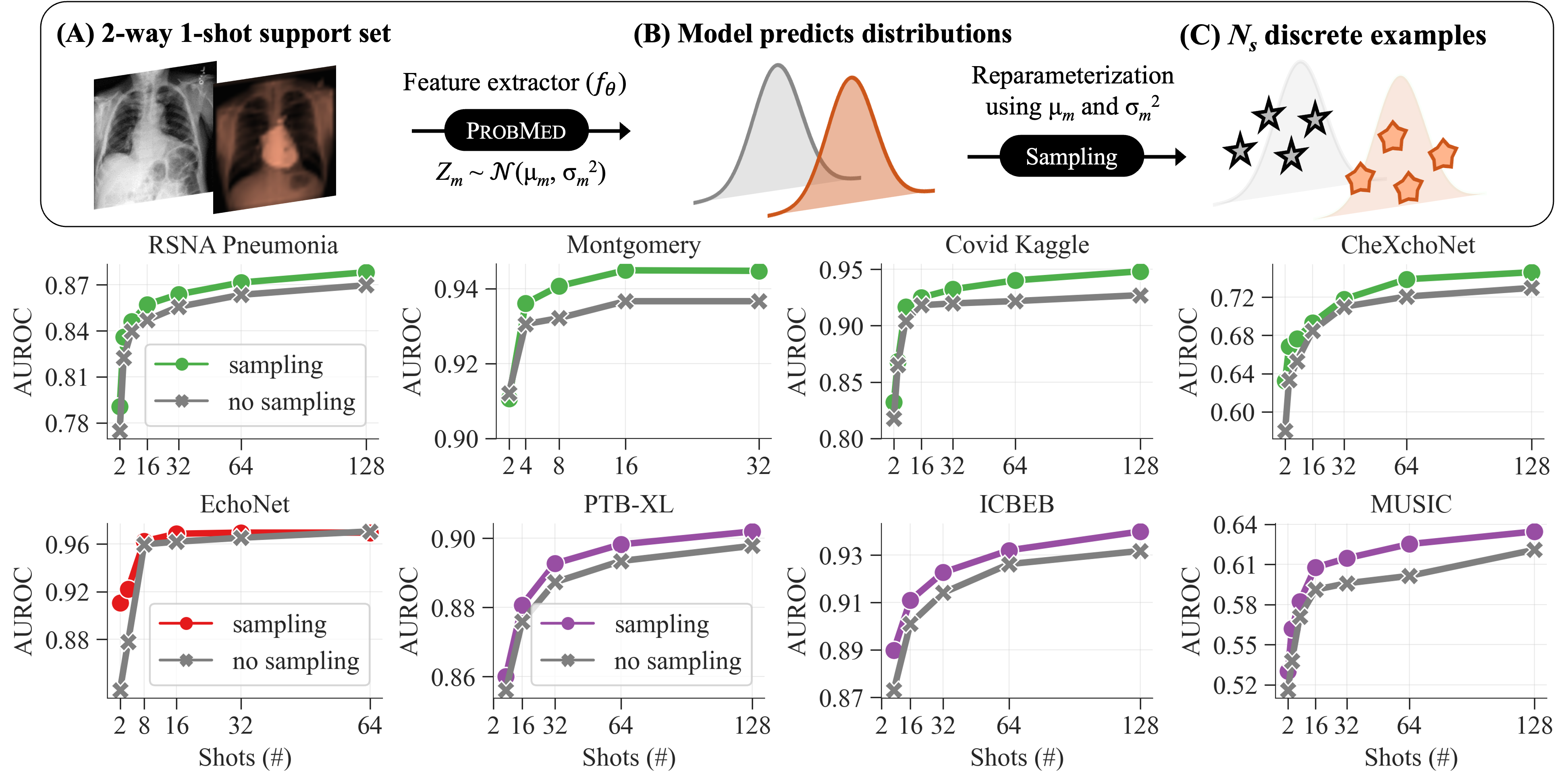}
    \caption{Probabilistic sampling in \textsc{ProbMED} improves few-shot performance. \textbf{(A–C)} Our probabilistic feature sampling strategy. \textsc{ProbMED} extracts a probabilistic distribution \( Z \sim \mathcal{N}(\mu, \sigma^2)\) for inputs. We leverage the reparameterization trick to sample $16$ distinct feature embeddings, capturing latent uncertainty. \textbf{(Bottom)} Few-shot AUROC comparison across 8 datasets. Sampling-based embeddings (\textit{colored lines}) against using only the $\mu$ embedding (\textcolor{Gray}{\textit{grey lines}}). For reference, \textcolor{Gray}{\textit{grey lines}} is \textsc{ProbMED} results in Tab.~\ref{tab:cxrfewshot}, ~\ref{tab:ecgfewshot}, and~\ref{tab:echofewshot}. }
    \label{fig:samplingfewresults}
\end{figure*}

\noindent\textbf{Sampling to Improve Few-shot}: We explore a novel probabilistic FS learning approach that harnesses feature sampling to enrich representation in FS regimes--enabled through probabilistic modeling. Unlike methods that rely solely on deterministic embeddings, \textsc{ProbMED} models each input as a distribution (i.e., refer to $\S$\ref{sec:bindingmodal},~\cref{eq:defined_distribution}. Let $\mathcal{P}(z)$ denote the learned embedding distribution for an input modality. Using the reparameterization trick, we drew $n$ distinct feature samples from $\mathcal{P}(z)$ to create additional data $\{z^{\prime}_j\} _ {j=1} ^ {n}$, effectively capturing the latent uncertainty of the data. When applied to our FS datasets with $n=16$, we improved over \textit{all our FS} results (i.e., $\S$\ref{sec:zerofewshot}) with this probabilistic sampling approach, as shown in~\cref{fig:samplingfewresults}. This sampling approach is also visualized in~\cref{fig:highlevel_samplingoverview}, where, in a 2-way 3-shot scenario, the decision boundary for traditional methods is not enough to approximate the true decision boundary. With sampling, the decision boundary is much closer. ~\cref{fig:highlevel_samplingoverview} and~\cref{fig:samplingfewresults} illustrate our approach, which we believe could be useful for limited data availability.
\section{Ablation}
We performed ablations to evaluate the impact of our \textsc{ProbMED} design choices, concentrating on the similarity metric applied for probabilistic embeddings and additional enhancements such as our SIS loss and BN. ~\cref{tab:distance} compares four similarity metrics: cosine, CSD, Bhattacharyya, and Hellinger, on MIMIC-CXR, MIMIC-ECG, and MIMIC-ECHO (that is, all possible pairs of modality-text). Cosine represents the \textbf{deterministic} training approach on these modality pairs and performs well in some recall tests; however, Hellinger consistently achieved the highest Recall@1 and Recall@5 scores across datasets, confirming its effectiveness. Hellinger offers a more expressive similarity measure by accounting for the overlap between probability distributions. This enables the model to capture more nuanced cross-modality relationships. The low performance of CSD in~\cref{tab:distance} arises from the limitations of its metric, not probabilistic training, and shows that a robust probabilistic distance (e.g., Hellinger) is needed to outperform deterministic baselines.
\begin{table}[ht]
\centering
\begin{subtable}[t]{\columnwidth}
\caption{\textbf{Ablation training probabilistic similarity metric.}}
\label{tab:distance}
\centering
\resizebox{\columnwidth}{!}{%
\begin{tabular}{l?cc?cc?cc?c}
\thickhline
\multirow{2}{*}{Similarities} & \multicolumn{2}{c?}{\textcolor{Green}{MIMIC-CXR}} & \multicolumn{2}{c?}{\textcolor{Purple}{MIMIC-ECG}} & \multicolumn{2}{c?}{\textcolor{Maroon}{MIMIC-ECHO}} & \multirow{2}{*}{RSUM}\\
                        & R@1   & R@5   & R@1   & R@5   & R@1   & R@5 & \\
\hline
Cosine$^*$ & 42.4 & \underline{64.5}& \textbf{48.9} & 87.3 & \textbf{2.0} & \underline{6.3} & \underline{251.4} \\
CSD      & 32.1  & 50.6  & 40.9  & 54.8  & \underline{1.0}  & 5.0 & 184.4 \\
Bhattacharyya      & \underline{43.4}  & 63.7  & 41.9  & \underline{87.9}  & \underline{1.0}  & 3.2 & 241.1\\
Hellinger   & \textbf{44.5}    & \textbf{67.7}    & \underline{48.1}    & \textbf{91.1}    & \textbf{2.0}    & \textbf{6.9} & \textbf{260.3} \\
\thickhline
\end{tabular}%
}

\end{subtable}

\vspace{0.2em}

\begin{subtable}[t]{\columnwidth}
\caption{\textbf{Ablation of additional loss and batch normalization.}}
\label{tab:addition}
\centering
\resizebox{\columnwidth}{!}{%
\begin{tabular}{l?cc?cc?cc?c}
\thickhline
\multirow{2}{*}{Method} & \multicolumn{2}{c?}{\textcolor{Green}{MIMIC-CXR}} & \multicolumn{2}{c?}{\textcolor{Purple}{MIMIC-ECG}} & \multicolumn{2}{c?}{\textcolor{Maroon}{MIMIC-ECHO}} & \multirow{2}{*}{RSUM}\\
                        & R@1   & R@5   & R@1   & R@5   & R@1   & R@5 & \\
\hline
Hellinger & 44.5  & 67.7  & \underline{48.1}  & \textbf{91.1}  & 2.0  & 6.9 & 260.3 \\
+ SIS Loss & 46.2  & \textbf{71.5}  & \underline{48.1}  & \underline{87.0}  & \underline{2.1}  & 6.4 & 261.3 \\
+ BatchNorm & \underline{47.2} & 70.7 & 47.9 & 86.7 & 2.3 & \underline{7.5} & \underline{262.3} \\
\hline
+ Both (Ours) & \textbf{47.9}    & \underline{71.4}  & \textbf{48.3}    & \underline{87.0}  & \textbf{2.4}    & \textbf{7.8} & \textbf{264.8} \\
\thickhline
\end{tabular}%
}

\end{subtable}

\caption{(a) Similarity metric performance across \textcolor{Green}{MIMIC-CXR}, \textcolor{Purple}{MIMIC-ECG}, and \textcolor{Maroon}{MIMIC-ECHO}. (b) Results show the impact of the new loss and batch normalization on performance.$^*$$\mu$ embedding was used to calculate the cosine similarity.}
\label{tab:ablation}
\end{table}

~\cref{tab:addition} examines two enhancements: our SIS loss integrated into pretraining and BN applied before feature aggregation. Using the SIS loss on top of the Hellinger baseline improved performance in MIMIC-CXR, while BN further boosted scores in MIMIC-ECHO. Combined, we saw the highest Recall@1 and Recall@5 across datasets, underscoring the complementary benefits of modeling through sampling (SIS loss) and stabilizing feature distributions (BN).

\section{Conclusion}
\textbf{\textsc{ProbMED}} is a probabilistic framework for binding multimodal medical data—modeling each modality as a distribution to capture many-to-many relationships—and outperforms existing Med-VLPMs in retrieval, ZS, and FS across 13 medical datasets. Future works include integrating additional modalities, utilizing label efficient fine-tuning, and exploring more probabilistic use cases. In summary, our probabilistic approach to multimodal Med-VLPM introduce fresh perspectives in the field.

\section*{Acknowledgments}
Study was funded by NSERC RGPIN-2022-05117. CM holds the Chair in Medical Imaging at the Joint Department of Medical Imaging at University Health Network and University of Toronto (UofT). YG holds CIHR Canada Graduate Scholarship - Doctoral. SK is funded by the doctoral fellowship from Data Sciences Institute at UofT.

{
    \small
    \bibliographystyle{ieeenat_fullname}

}

\clearpage
\appendix
\setcounter{page}{1}
\maketitlesupplementary
\begin{table*}[t]
  \centering
  \resizebox{0.95\textwidth}{!}{
  \begin{tabular}{l|c|c|cccccc}
    \thickhline
    \textbf{Dataset} & \textbf{Pretrain?} & \textbf{Modalities} & \textbf{Task} & \textbf{\#Cls} & \textbf{\#train} & \textbf{\#valid} & \textbf{\#test} \\ 
    \hline
    \textcolor{Green}{MIMIC-CXR}~\cite{johnson2019mimiccxr} & \cmark & CXR + TXT & Pretrain/Retrieval/\textbf{Multimodal-Classification} & 12 & 86,853 & 12,059 & 24,799 \\
    \textcolor{Green}{OpenI}~\cite{demner2016preparing_openi} & \textcolor{Gray}{\xmark} & CXR + TXT & Retrieval & - & - & - & 2,864 \\
    \textcolor{Green}{CheXpert5x200}~\cite{irvin2019chexpert} & \textcolor{Gray}{\xmark} & CXR + TXT & Retrieval & 5 & - & - & 1,000 \\
    \textcolor{Green}{RSNA}~\cite{shih2019augmenting_rsna} & \textcolor{Gray}{\xmark} & CXR & Classification & 2 & 18,678 & - & 5,338 \\
    \textcolor{Green}{COVID}~\cite{chowdhury2020can_kaggle_covid1} & \textcolor{Gray}{\xmark} & CXR & Classification & 2 & 11,028 & - & 2,780 \\
    \textcolor{Green}{Montgomery}~\cite{candemir2013lungmontgomery,jaeger2013automaticmontgomery2} & \textcolor{Gray}{\xmark} & CXR & Classification & 2 & 32 & - & 106 \\
    \textcolor{Green}{CheXchoNet}~\cite{bhave2024deepchexcho} & \textcolor{Gray}{\xmark} & CXR$^*$ & Classification & 2 & 64,619 & 3,303 & 3,667 \\
    \hline
    \textcolor{Purple}{MIMIC-ECG}~\cite{gow_mimicecg} & \cmark & ECG + TXT & Pretrain/Retrieval/\textbf{Multimodal-Classification} & - & 88,291 & 12,065 & 24,644 \\
    \textcolor{Purple}{PTB-XL}~\cite{strodthoff2020deep_ptbxl} & \textcolor{Gray}{\xmark} & ECG + TXT & Retrieval/Classification & 71 & 17,415 & 2,183 & 2,198 \\
    \textcolor{Purple}{ICBEB}~\cite{liu2018open_icbeb} & \textcolor{Gray}{\xmark} & ECG & Classification & 9 & 5,501 & - & 1,376 \\
    \textcolor{Purple}{MUSIC}~\cite{martinmusic} & \textcolor{Gray}{\xmark} & ECG$^*$ & Classification & 2 & 512 & - & 125 \\
    \hline
    \textcolor{Maroon}{MIMIC-ECHO}~\cite{gow2023mimicecho} & \cmark & ECHO + TXT & Pretrain/Retrieval & - & 13,732 & 3,880 & 1,957 \\
    \textcolor{Maroon}{EchoNet-Dynamic}~\cite{ouyang2019echonet} & \textcolor{Gray}{\xmark} & ECHO & Classification & 2 & 7,394 & 1,273 & 1,264 \\
    \hline
    MIMIC-CONNECT~\cite{gow_mimicecg,johnson2019mimiccxr} & \cmark & ECG + CXR & \textbf{Multimodal-Classification} & - & 22,397 & 3,292 & 6,664 \\
    \thickhline
  \end{tabular}}
  \caption{\textbf{ALL} datasets for \textcolor{Green}{CXR}, \textcolor{Purple}{ECG}, and \textcolor{Maroon}{ECHO} modalities. Pretrain column represents data used to train ProbMED. Modalities highlight the modality types. $^*$These datasets contain corresponding modality and \textbf{ECHO}-based labels derived from the ECHO-report.}
  \label{tab:full_dataset}
\end{table*}

\section{Datasets}
\label{suppsec:datasets}
We pretrained \textsc{ProbMED} exclusively on MIMIC datasets, specifically MIMIC-CXR~\cite{johnson2019mimiccxr}, MIMIC-ECG~\cite{gow_mimicecg}, and MIMIC-ECHO~\cite{gow2023mimicecho}. Additionally, we constructed the \textbf{MIMIC-CONNECT} subset (following \cite{gao2024medbind}), which contains patients with both CXR and ECG data (details in $\S$\ref{subsec:mimicconnect}). \textbf{We maintained the same patient-level differences in all MIMIC datasets to prevent data contamination.} Furthermore, MIMIC-CONNECT enabled the evaluation of a multimodal approach, utilizing CXR and ECG for prediction. The details are described below and~\cref{tab:full_dataset}. We split up the datasets used by their respective modality. \par
\subsection{Chest X-ray datasets}
\noindent\textcolor{Green}{\textbf{MIMIC-CXR~\cite{johnson2019mimiccxr}.}} This dataset is used for both pretraining and evaluation. It consists of CXR with their paired radiology reports. We preprocessed each case’s CXR and corresponding text using methods outlined in~\cite{you2023cxrclip}. We restricted the dataset to only the Anterior-Posterior (AP and PA) views. For pretraining, we used a predefined training split of the dataset for CXR-text binding. \par
\noindent\textcolor{Green}{\textbf{Kaggle COVID~\cite{chowdhury2020can_kaggle_covid1}.}} This public dataset comprises CXR images annotated with binary COVID-19 labels and was used solely for ZS and FS task. For ZS testing, we generated prompts as suggested in~\cite{you2023cxrclip}. \par
\noindent\textcolor{Green}{\textbf{RSNA Pneumonia~\cite{shih2019augmenting_rsna}.}} The RSNA Pneumonia dataset contains CXR images of pneumonia cases and is publicly available through the National Institutes of Health database. This dataset was also only used for ZS and FS task. For ZS classification, we used a prompt, \texttt{"Chest X-ray findings consistent with lung infection"} to explain pneumonia for ZS evaluation. We followed~\cite{gao2024medbind} for splitting the dataset into train and test set. \par
\noindent\textcolor{Green}{\textbf{Montgomery~\cite{candemir2013lungmontgomery,jaeger2013automaticmontgomery2}.}} This consists of CXR images collected from Tuberculosis Control program in Montgomery County, Maryland. It includes annotations for tuberculosis and other thoracic abnormalities, providing a challenging evaluation subset. For consistency with our methodology, we used the same prompt for the RSNA Pneumonia dataset to explain tuberculosis for ZS evaluation. \par
\noindent\textcolor{Green}{\textbf{OpenI~\cite{demner2016preparing_openi}.}} This dataset comprises CXR images, corresponding radiology reports, and clinical findings extracted from the Indiana University hospital database. Among all CXR images, we took the CXRs with the AP view. We used this dataset for TEXT-to-CXR retrieval evaluation. \par
\noindent\textcolor{Green}{\textbf{Chexpert5x200~\cite{irvin2019chexpert}.}} Following the formulation in previous works~\cite{you2023cxrclip, gao2024medbind, huang2021gloria}, CheXpert5x200 is a multi-class classification subset derived from CheXpert-1.0~\cite{irvin2019chexpert}. It consists of five distinct classes (i.e., atelectasis, cardiomegaly, consolidation, edema, and pleural effusion), with 200 images per class. We followed~\cite{huang2021gloria} for the test split. This dataset's experimental setup (TEXT-to-CXR retrieval) and evaluation metrics mirrored those applied to the other datasets. \par

\noindent\textcolor{Green}{\textbf{CheXchoNet~\cite{bhave2024deepchexcho}.}} This open-public CXR dataset has unique pairs of CXR with gold-standard ECHO labels. We used a label of composite of severe left ventricular hypertrophy and dilated left ventricle, which are both significant findings that can be found on ECHO and also in CXR~\cite{bhave2024deepchexcho}. In our experiments, CheXchoNet was utilized primarily for evaluation and acted as an emergent alignment dataset (i.e., detecting unseen diseases during the evaluation). ZS and FS settings assessed the model’s performance in identifying ECHO-based pathologies. \par
\subsection{ECG datasets}
\noindent\textcolor{Purple}{\textbf{MIMIC-ECG~\cite{gow_mimicecg}.}} The dataset comprises 10-second, 12-lead ECG recordings originally sampled at 500 Hz. These signals were down-sampled to 100 Hz using a low-pass filter to reduce noise and computational overhead~\cite{christov2017pseudo}. Each ECG is accompanied by machine-generated reports and links \texttt{cart\_id} to free-form textual data. When available, the free-form text was used to generate corresponding ECG descriptions; otherwise, the machine reports were employed. This dataset played a dual role in our study, as it was used for training and evaluation. \par

\noindent\textcolor{Purple}{\textbf{PTB-XL~\cite{strodthoff2020deep_ptbxl}.}} This ECG dataset is a large-scale, publicly available repository of 12-lead ECG recordings and contains annotated ECGs with comprehensive diagnostic labels and expert assessments. For our experiments, we applied similar preprocessing steps as with MIMIC-ECG, including downsampling to 100 Hz using an appropriate low-pass filter. The dataset was used solely for evaluation: retrieval, ZS, and FS settings. \par

\noindent\textcolor{Purple}{\textbf{ICBEB~\cite{liu2018open_icbeb}.}} This dataset provides 12-lead ECG recordings, annotated with diagnostic information. We applied consistent pre-processing, that is, low-pass filtering and downsampling to 100 Hz, to ensure compatibility with our training protocols. The ICBEB dataset was used exclusively for evaluation with ZS and FS experiments.\par
\noindent\textcolor{Purple}{\textbf{MUSIC~\cite{martinmusic}.}} This ECG dataset comprises vectorcardiograms (VCGs) originally sampled at 1000 Hz. Each VCG recording is paired with ECHO labels indicating key clinical parameters: SLVH, DLV, and LVEF. Although the ECHO modality is not provided, the associated labels offer valuable diagnostic insights. To use this dataset, we applied the Kors regression transformation~\cite{vondrak2022statisticalkors} to convert the VCGs into 12-lead ECGs, which were subsequently down-sampled to 100 Hz. MUSIC was used solely for evaluation in our emergent ZS and FS experiments. Like CheXchoNet, we used a label for a composite of severe left ventricular hypertrophy and dilated left ventricle.\par
\subsection{ECHO datasets}
\noindent\textcolor{Maroon}{\textbf{MIMIC-ECHO~\cite{gow2023mimicecho}.}} This dataset consists of ECHOs from various patients in the MIMIC dataset cohort. We matched patients using \texttt{hadm\_id} to connect data to discharge notes in MIMIC-IV~\cite{johnson2023mimic4}. In this study, we used \textbf{all ECHOs connected to a discharge note containing ECHO-related text}. These texts were first processed by Llama3.1-Instruct 8B~\cite{dubey2024llama}, followed by manual verification by human experts. DICOMs for each ECHO were processed into individual ECHOs. All frames were used as augmentations during training, while only the first frame was used for evaluation.\par

\noindent\textcolor{Maroon}{\textbf{EchoNet-Dynamic~\cite{ouyang2019echonet}.}} We used this ECHO dataset comprising apical-4-chamber ECHO videos and corresponding left ventricular ejection fraction (LVEF) labels. Continuous LVEF values were constructed to be binary with threshold LVEF$<$40\%, based on~\cite{vasan2018epidemiology}. Each video has been preprocessed to a standardized ($3\times112\times112$). A single frame corresponding to the end-systolic (ES) phase of the left ventricle was extracted from the ECHO video, as labeled in the EchoNet-Dynamic dataset. We extrapolated the frame into ($3\times224\times224$) resolution using cubic interpolation to match the resolution with MIMIC-ECHO and used this dataset for evaluation in ZS and FS.\par
\subsection{MIMIC-CONNECT}
\label{subsec:mimicconnect}
\textbf{MIMIC-CONNECT}. We derived this dataset by linking MIMIC-CXR and MIMIC-ECG to MIMIC-IV~\cite{johnson2023mimic4}. We matched \texttt{subject\_id} and \texttt{hadm\_id} ensuring that the modality recording times were within 7-day window. For cases with available visit identifiers, \texttt{hadm\_id}, we directly paired MIMIC-CXR and MIMIC-ECG to form this dataset that we refer to as MIMIC-CONNECT; when \texttt{hadm\_id} was unavailable, the pairing was based solely on \texttt{subject\_id} and a 7-day window between the CXR and ECG recordings. This dataset was used for training and evaluation--multimodal ZS and FS classification tasks.\par
\subsection{Data Representation}
We used conventional representations for each modality. CXR images, initially single-channel ($1\times224\times224$), are duplicated across channels to match the standard 3-channel (RGB) inputs, making them into the size of ($3\times224\times224$). ECG signals are treated as 12 distinct leads over time, producing a ($12\times1000$) tensor for a 10-second recording sampled at 100 Hz. Processing ECHO videos follows the method in ECHO-CLIP~\cite{christensen2024visionechoclip}, utilizing separate frames with ($3\times224\times224$) resolution instead of the entire video as an input. Finally, text data is tokenized with a BERT-based tokenizer~\cite{devlin2018bert}, with sequences padded or truncated to 100 tokens to fit typical medical report lengths.

\section{Loss Function Implementation}
\subsection{Hellinger Loss Calculation}
\label{subsec:hellingercalc}
The Hellinger equation, from~\cite{pardo2018statistical}, is calculated in this paper based on the following simplifications--aligned with the propositions in the paper. First, we bring the \cref{eq:closedform_hellinger} from the main manuscript here (as a reference):
\[
\begin{split}
&H^2(q_n,k_t)= 1 - \frac{\det(\boldsymbol{\Sigma}_n)^{\frac{1}{4}}\det(\boldsymbol{\Sigma}_t)^{\frac{1}{4}}}{
\det\!\Bigl(\frac{\boldsymbol{\Sigma}_n+\boldsymbol{\Sigma}_t}{2}\Bigr)^{\frac{1}{2}}}
\\&
\quad\times \exp\Bigr((-\frac{1}{8}(\mu_n-\mu_t)^\top 
\Bigl(\frac{\boldsymbol{\Sigma}_n+\boldsymbol{\Sigma}_t}{2}\Bigr)^{-1} (\mu_n-\mu_t)\Bigr).
\end{split}
\]
\noindent Assuming that the covariance matrices are diagonal, i.e., our base assumption in $\S$~\ref{subsection:preliminaries}.  The determinants can be written as:
\begin{equation}
\label{seq:productdeterminant}
    \det(\boldsymbol{\boldsymbol{\Sigma}}_n) = \prod_{o=1}^D \sigma_{n,o}^2 \quad \text{and} \quad \det(\boldsymbol{\boldsymbol{\Sigma}}_t) = \prod_{o=1}^D \sigma_{t,o}^2.
\end{equation}
From \cref{seq:productdeterminant}, we see that the product term in \cref{eq:closedform_hellinger} can be expressed as:
\begin{equation}
\label{seq:part1_hell}
\begin{split}
    \det(\boldsymbol{\Sigma}_n)^{\frac{1}{4}}\det(\boldsymbol{\Sigma}_t)^{\frac{1}{4}} &= \prod_{o=1}^{D} (\sigma_{n,o}^2)^{\frac{1}{4}} (\sigma_{t,o}^2)^{\frac{1}{4}} \\
    & = \prod_{o=1}^{D} (\sigma_{n,o}\sigma_{t,o})^{\frac{1}{2}},
\end{split}
\end{equation}
\noindent Next, the determinant of the average term is:
\begin{equation}
\label{seq:part2_hell}
\det\!\Bigl(\frac{\boldsymbol{\Sigma}_n+\boldsymbol{\Sigma}_t}{2}\Bigr)^{\frac{1}{2}} = \prod_{o=1}^{D} \left(\frac{\sigma_{n,o}^2+\sigma_{t,o}^2}{2}\right)^{\frac{1}{2}}.
\end{equation}
\noindent Furthermore, because the matrices are diagonal, the quadratic form in the exponent simplifies to a sum over dimensions:
\begin{equation}
\begin{split}  
(\mu_n-\mu_t)^\top & \Bigl(\frac{\boldsymbol{\Sigma}_n+\boldsymbol{\Sigma}_t}{2}\Bigr)^{-1} (\mu_n-\mu_t) \\ & = \sum_{o=1}^D \frac{(\mu_{n,o}-\mu_{t,o})^2}{\frac{\sigma_{n,o}^2+\sigma_{t,o}^2}{2}} \\&= 2\sum_{o=1}^D \frac{(\mu_{n,o}-\mu_{t,o})^2}{\sigma_{n,o}^2+\sigma_{t,o}^2}.
\end{split}
\end{equation}
Thus, given all of this, and $\exp$ properties. We arrived at the formulation:
\begin{equation}
\begin{split}
& H^2(q_n,k_t) = 1 - \\& \prod_{o=1}^{D} \left[ \frac{({\sigma_{n,o}\sigma_{t,o}})^{\frac{1}{2}}}{{(\frac{\sigma_{n,o}^2+\sigma_{t,o}^2}{2}})^{\frac{1}{2}}} \exp\!\left(-\frac{(\mu_{n,o}-\mu_{t,o})^2}{4(\sigma_{n,o}^2+\sigma_{t,o}^2)}\right) \right].
\end{split}
\end{equation}

\noindent This can be simplied to in our formulation in \cref{eq:our_hellinger}:
\[
\begin{split}
&H^2(q_n,k_t) = 1 - \\& \prod_{o=1}^{D} \left[ \left(\frac{{2\sigma_{n,o}\sigma_{t,o}}}{{{\sigma_{n,o}^2+\sigma_{t,o}^2}}}\right)^{\frac{1}{2}} \exp\!\left(-\frac{(\mu_{n,o}-\mu_{t,o})^2}{4(\sigma_{n,o}^2+\sigma_{t,o}^2)}\right) \right].
\end{split}
\]

The pseudocode for calculating the Hellinger loss involves similar computations but uses the \textbf{logsumexp} trick (\cref{alg:helliniger}). Hellinger distance is bounded between 0 and 1, where zero is the \textbf{same distribution}, and one is \textbf{far apart}. To use this in losses, we instead use $1-\sqrt{H^2}$, where $H^2$ is the squared Hellinger distance. This lets us view our loss like \textbf{cosine similarity}. The implementation and pseudo-code are provided below.

\algrenewcommand{\algorithmiccomment}[1]{\hfill$\triangleright$ \textcolor{RoyalBlue}{#1}}
\algnewcommand{\LineComment}[1]{\State \(\triangleright\) \textcolor{RoyalBlue} {#1}}
\begin{algorithm}[H]
\caption{Hellinger Distance}
\label{alg:helliniger}
\begin{algorithmic}[1]
\Procedure{ComputeHellinger}{$q_n$, $k_t$}
    \LineComment {Get mean and log-variance:}
    \State \quad $ \mu_n, \log(\sigma^2_n) \gets q_n$
    \State \quad $ \mu_t, \log(\sigma^2_t) \gets k_t$
    \LineComment {Convert log-variance to variance:}
    \State \quad $ \sigma^2_n \gets \exp(\log\sigma^2_n))$
    \State \quad $ \sigma^2_t \gets \exp(\log\sigma^2_t))$
    \LineComment {Compute hellinger terms, logsumexp trick:}
    \State \quad $ \sigma_{\text{prod}} = \sqrt{\sigma^2_n\sigma^2_t}$ 
    \State \quad $ \sigma^2_{\text{sum}} = \sigma^2_n + \sigma^2_t$
    \State \quad $T_1 = \frac{1}{2}\times \log \left( \frac{2\times\sigma_{\text{prod}}}{\sigma^2_{\text{sum}}} \right)$
    \State \quad $T_2 = \frac{1}{4}\times \frac{(\mu_n - \mu_t)^2}{\sigma^2_{\text{sum}}}$
    \LineComment {Compute the sum of logs across \textit{D} dims:}
    \State \quad $ T = \sum(T_1 + T_2)$
    
    \LineComment {Convert back with exp:}
    \State \quad $ p = \exp(T)$
    
    \LineComment {Compute squared Hellinger distance:}
    \State \quad $ H^2 = 1 - p$
    \State \Return $\sqrt{H^2}$
\EndProcedure
\end{algorithmic}
\end{algorithm}

\begin{algorithm}[H]
\caption{SIS Loss Computation}
\label{alg:sampling_loss}
\begin{algorithmic}[1]
\Procedure{ComputeSISLoss}{$\mu$, $\log\sigma^2$, $N_s=2$}
    \State \Comment{Compute the standard deviation w/ log-variance:}
    \State $\sigma \gets \exp\left(0.5 \times \log\sigma^2\right)$
    \State $S \gets$ \textbf{empty list}
    \For{$l \gets 1$ \textbf{to} $N_s$}
        \State \Comment{Sample $\epsilon^l$ from standard normal distribution:}
        \State $\epsilon^l \gets \text{RandomNormal}(\text{shape}(\mu))$
        \State \Comment{Reparameterization is used to obtain samples:}
        \State $s \gets \mu + \operatorname{diag}({\sigma})\epsilon^l$
        \State Append $s$ to $S$
    \EndFor
    \State \Comment{Compute the 2N InfoNCE loss between samples:}
    \State $L \gets \text{InfoNCE}(S[0], S[1])$
    \State \Return $L$
\EndProcedure
\end{algorithmic}
\end{algorithm}
\subsection{Intra-modality Loss}
\label{sup:intramodality_loss}
The pseudocode in \cref{alg:sampling_loss} outlines our SIS loss computation for within-modality learning in three main steps. First, we calculate the standard deviation from the given $\log(\sigma^2)$ by taking the exponential of half the $\log(\sigma^2)$. Next, we sample a normal distribution's noise vector $\epsilon$. By applying the reparameterization trick, scaling this noise by the computed standard deviation, and shifting it by the mean, we generate a sample from the desired multivariate normal distribution with diagonal covariance~\cite{kingma2013autoVAE}. This process is repeated to produce the required samples, $N_s$ (here, $N_s = 2$ in our implementation--analagous to SimCLR~\cite{chen2020simplesimclr}).
\section{Model Architecture}
\label{sec:fullmodel} 
\textsc{ProbMED} was built on PCME++~\cite{chun2023improvedpcmepp}, which trains separate encoders for different data modalities (e.g., images and text) and represents each input as a normal distribution in a shared latent space. Specifically, each encoder outputs two D-dimensional vectors—one for $\mu$ and one for $\log(\sigma^2)$—that parameterize a Gaussian distribution. This setup allows the model to capture uncertainty and variability in the learned embeddings. Following PCME++, we found that using traditional encoders with two outputs effectively trained the probabilistic models. Extrapolated from PCME++, we produced $\mu$ and $\log (\sigma^2)$, using a duplicated final Transformer layer, i.e., the first branch is initialized with the same weights as the backbone (for $\mu$). In contrast, the second branch (for $\log (\sigma^2)$) is initialized randomly. We adopt the GPO~\cite{chen2021learningGPO} for feature aggregation, improving training stability and performance.

\begin{figure}
    \centering
    \includegraphics[width=\columnwidth]{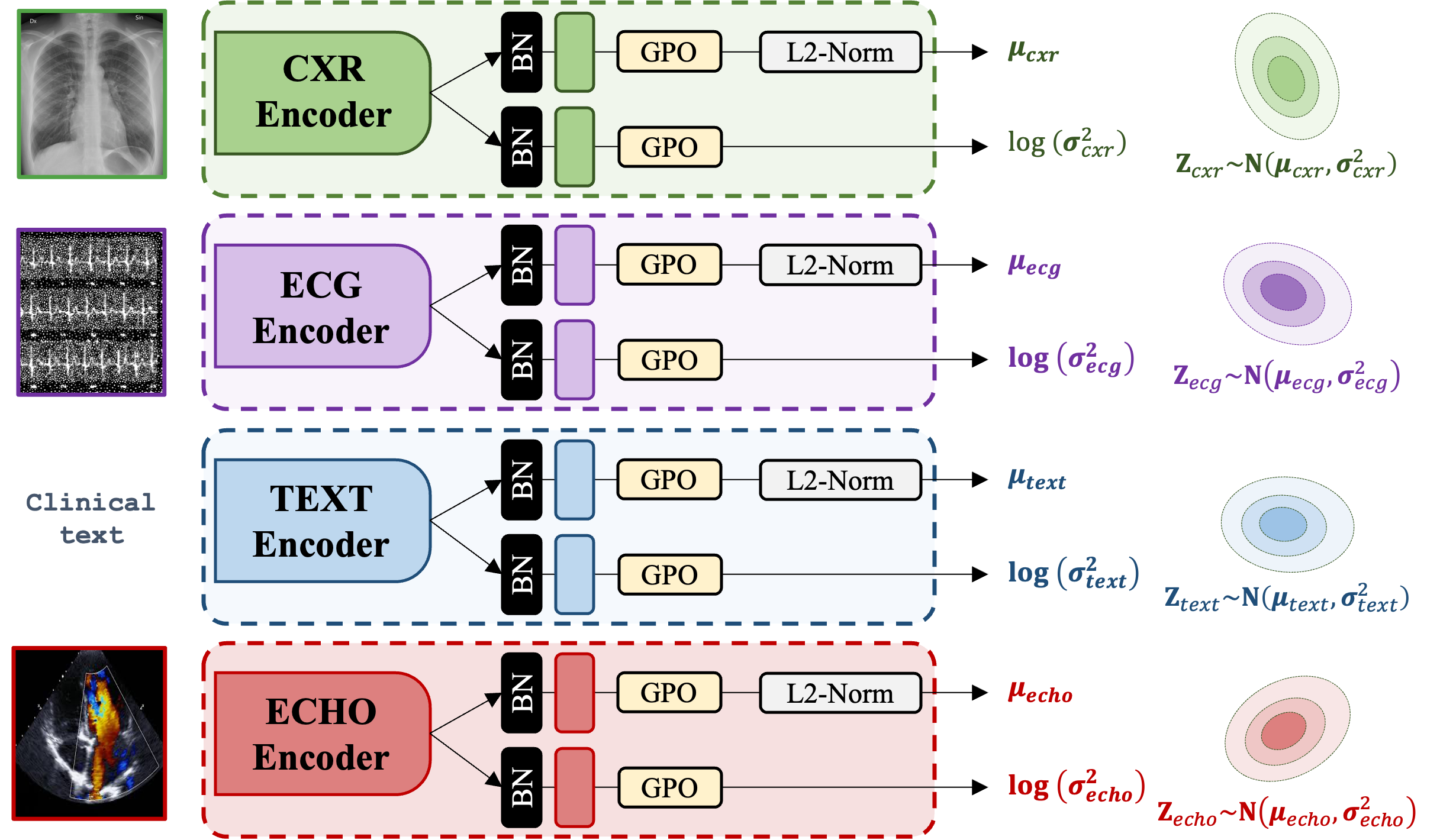}
    \caption{\textsc{ProbMED} model architecture. Encoders follow standard models. The proposed method of extracting the $\mu$ and $\log(\sigma^2)$ follows PCME++~\cite{chun2023improvedpcmepp}. BN represents BatchNorm1d and GPO is the Generalized Pooling Operator~\cite{chen2021learningGPO}.}
    \label{sfig:modelarchitecture}
\end{figure}

~\cref{sfig:modelarchitecture} shows an overview of our model architecture. Our framework differs from the original PCME++ in two key ways. First, we introduce batch normalization layers following the encoders to normalize the input based on the mini-batch mean and variance. Second, we extend the architecture to simultaneously handle multiple medical modalities—such as CXR, ECG, text, and ECHO—by adding separate encoder branches for each modality. These multimodal embeddings are learned in a unified latent space, facilitating cross-modal alignment and downstream clinical tasks.

\section{Pretraining details}
\label{suppsec:pretraining}
In \textsc{ProbMED}, each modality is processed by a dedicated encoder chosen for its domain-specific strengths. Inspired by PCME++—highlighting the importance of modality-specific representations when transitioning to a probabilistic embedding space—we adopt state-of-the-art pre-trained models where they are most effective~\cite{chun2023improvedpcmepp}. Our text data is encoded using BioBERT~\cite{lee2020biobert} to capture rich, domain-aware linguistic nuances, while the CXR modality benefits from the robust feature extraction of the Swin-tiny~\cite{liu2021swin, huggingface2023swintiny} model. For ECHO, we employ ConvNeXt~\cite{liu2022convnetconvnext} CLIP, with ECHO-CLIP weights~\cite{christensen2024visionechoclip} to effectively model its complex visual patterns. In contrast, our ECG encoder is built on a streamlined ResNet1D architecture and trained from scratch, as our experiments did not reveal any advantages from pretraining for this modality. This modular design enables \textsc{ProbMED} to leverage the strengths of specialized encoders within a unified framework for cross-modal probabilistic learning.

When we pretrain \textsc{ProbMED}, we utilized data augmentations for the input modalities. For CXR, we applied the random cropping to 224 $\times$ 224 from 256 $\times$ 256, horizontal flipping, color jittering, and random affine transformations following~\cite{wang2022medclip}. For ECG data augmentation, we applied adding random Gaussian noise. For the ECHO data, we applied randomized color jittering, gray scaling, and adding random Gaussian noise.

We trained \textsc{ProbMED} with the following hyperparameters detailed in~\cref{stab:model_config}. The hyperparameters for \textbf{the final loss} (\cref{eq:final_loss}) are $\alpha = 1.0$, $\beta = 0.5$, $\gamma = 0.0001$. We set the temperature scale with $\tau = 0.07$ for all related losses; we explore more $\tau$ parameters in $\S$\ref{suppsec:ablations}. All our experiments (including pretraining) were conducted on a single 48GB L40S or a 40GB A100 GPU.
\begin{table*}[t]

    \centering
\resizebox{0.66\textwidth}{!}{%
    \begin{tabular}{l?cccc}
        \thickhline
        \textbf{Config} & \textcolor{RoyalBlue}{\textbf{TEXT}} & \textcolor{Green}{\textbf{CXR}} & \textcolor{Purple}{\textbf{ECG}} & \textcolor{Maroon}{\textbf{ECHO}} \\
        \hline
        Pretrained Models   & BioBERT~\cite{lee2020biobert}& Swin-tiny~\cite{liu2021swin} & XResNet-1d~\cite{he2019bagxresnet} & ECHO-CLIP~\cite{christensen2024visionechoclip} \\
        Final output dim.              & \multicolumn{4}{c}{$\mu$-output=512,  $\log(\sigma^2)$-output=512} \\
 
        Optimizer           & \multicolumn{4}{c}{AdamW} \\
        Optimizer Momentum  &\multicolumn{4}{c}{$\beta_1=0.9$,$\beta_2=0.95$} \\
        Learning Rate (LR)  & \multicolumn{4}{c}{1.00e-04} \\
        LR Scheduler        & \multicolumn{4}{c}{CosineAnnealingLR} \\
        Gradient clipping   & \multicolumn{4}{c}{1.0} \\
        Weight Decay        & \multicolumn{4}{c}{1.00e-05} \\
        Batch size          & \multicolumn{4}{c}{192} \\
        Total Epoches       & \multicolumn{4}{c}{120} \\
        \thickhline
    \end{tabular}
    }
\caption{\textsc{ProbMED} hyperparameters.}
\label{stab:model_config}
\end{table*}

\section{Evaluation Details}
In this section, we expand on the results presented in the main paper.
\subsection{Zero-Shot Prompts}
We generated dataset-specific prompts per label with varying descriptions of medical findings. Especially for the emergent ZS tasks (e.g., predicting the labels observable in ECHO using CXR), we generated 10 different prompts for the positive label since the emergent ability of the model often lacks capturing the meaning of the disease with using a single prompt. For non-emergent datasets, we chose using a simple prompt for the ZS task for simplicity. To keep the prompts concise and clinically relevant, we utilized short sentences focusing on the presence or absence of disease. We observed that succinct prompts improved interpretability and maintained performance in our ZS evaluations. For generating 80 different prompts used in~\cref{tab:zerofilter}, we used GPT-4o to paraphrase the baseline prompt making the descriptions of all prompts consistent. We compute the cosine distance between the text embeddings for each label (i.e., positive and negative label) and image embedding for ZS classification. For multiple prompts, we averaged the text embeddings for each label to generate a prototype, representing each label.\par
\subsection{Few-Shot Evaluation}
The FS results in \cref{stab:cxrfewshot_extended},~\cref{stab:ecgfewshot_extended}, and~\cref{stab:echofewshot_extended} used a traditional linear probing set-up~\cite{radford2021learning_clip}. We sampled $k$ training samples per class, where $k\in$ \{4, 16\}. These were chosen to highlight the use cases of our model in FS learning. Extended FS results (i.e., $k\in$ \{2, 4, 8, 16\}) are presented in the following subsection.

\section{Extended Results}
This section highlights the full results in many of the tables presented in the main text.

\subsection{Multimodal classification using MIMIC}
We showed that Top-K retrieval analysis identifies the most effective distance metric for retrieval in the main manuscript under ~\cref{tab:cxr_recall},~\cref{tab:ecg_recall}, and~\cref{tab:echo_recall}. Deterministic methods leverage cosine similarity distance, whereas probabilistic methods employ the distance measure used during model training (e.g., Hellinger distance for \textsc{ProbMED}). While this approach yielded strong results, we also emphasized evaluating all models under a \textbf{consistent} distance metric. Specifically, we adopted \textbf{cosine similarity} as the standard measure, which, in the case of probabilistic models, involves computing distances using \textbf{only the $\mu$ embedding}, as initially proposed in~\cite{chun2023improvedpcmepp}. The corresponding retrieval performance for various modality-text pairs is presented in \cref{stab:cosine_cxr_retrieval},~\cref{stab:cosine_ecg_recall},~\cref{stab:cosine_echo_recall}.\par
\begin{table}[t]
    \centering
    \label{stab:cosine_retrieval}    
\begin{subtable}[t]{\columnwidth} 
    \centering
    \caption{\textbf{Cosine-based TEXT-to-CXR retrieval}}
    \label{stab:cosine_cxr_retrieval}
    \resizebox{\columnwidth}{!}{ 
    \begin{tabular}{l?cc?cc?cc?c}
        \toprule
        \multirow{2}{*}{\textcolor{Green}{ }} 
        &\multicolumn{2}{c?}{\textcolor{Green}{MIMIC-CXR}} 
        & \multicolumn{2}{c?}{\textcolor{Green}{OpenI}}
        & \multicolumn{2}{c?}{\textcolor{Green}{Chexpert5x200}} 
        & \multirow{2}{*}{\textcolor{Green}{RSUM}} \\ 
         & R@1 & R@5 & R@1 & R@5 & R@1 & R@5 \\  
        \hline
      MedCLIP~\cite{wang2022medclip}  & 1.0 & 4.3 & 0.6 & 2.8 & 2.6 & 3.0 & 14.3\\
    CXR-CLIP~\cite{you2023cxrclip} & {\textbf{47.3}} & \underline{70.4} & \underline{12.7} & 25.2 & 8.5 & 23.0 & \underline{187.1} \\
    BiomedCLIP~\cite{zhang2023biomedclip} &  36.2 & 59.9 & 9.0 & 19.9 & 6.4 & 19.8 & 151.2 \\
    CheXzero~\cite{tiu2022expertchexzero}  & 26.7 & 50.0 & 5.8 & 15.1 & 3.5 & 17.8 & 118.9\\
    MEDBind~\cite{gao2024medbind}  & 40.8 & 67.5 & 11.6 & \underline{25.5} & 7.9 & 21.4 & 174.7 \\
            
            BioVil-T~\cite{bannur2023learningbiovil} & 28.4 & 58.2 & 8.1 & 18.9 & 4.9 & 17.1 & 135.6\\
            SAT~\cite{liu2023improvingSAT} & 40.3 & 69.2 & 6.7 & 14.7 & \textbf{9.1} & \textbf{26.7} & 166.7 \\
        \hline
        PCME++~\cite{chun2023improvedpcmepp}    & 4.5 & 21.9 & 9.5 & 20.6 & 1.3 & 4.6 & 62.4 \\
        \textsc{ProbMED} (Ours)    & \underline{47.0} & \textbf{70.8} & \textbf{13.2} & \textbf{28.1} & \underline{8.8} & \underline{23.9} & \textbf{191.8}\\
        \bottomrule
    \end{tabular}}
\end{subtable}
\begin{subtable}[t]{0.9\columnwidth} 
    \centering
    \caption{\textbf{Cosine-based TEXT-to-ECG retrieval}}
    \label{stab:cosine_ecg_recall}
    \resizebox{\columnwidth}{!}{ 
    \begin{tabular}{l?cc?cc?c}
        \toprule
        \multirow{2}{*}{\textcolor{Purple}{ }} 
        & \multicolumn{2}{c?}{\textcolor{Purple}{MIMIC-ECG}} 
        & \multicolumn{2}{c?}{\textcolor{Purple}{PTB-XL}} 
        & \multirow{2}{*}{\textcolor{Purple}{RSUM}} \\ 
        & R@1 & R@5 & R@1 & R@5 \\  
        \hline
        ECG-CLIP~\cite{gao2024medbind}  & 40.8 & 76.7 & 2.3 & 9.8 & 129.6\\
        MEDBind~\cite{gao2024medbind}  & \underline{44.1} & \underline{78.2} & \textbf{3.1} & \underline{12.1} & \underline{137.5}\\
        \hline
        PCME++~\cite{chun2023improvedpcmepp}  & 7.1 & 14.1 & 1.5 & 11.2 & 33.9\\
        \textsc{ProbMED} (Ours)  & \textbf{51.3} & \textbf{86.9} & \underline{2.4} & \textbf{12.7} & \textbf{153.3}\\
        \bottomrule
    \end{tabular}}
\end{subtable}
    \hspace{5pt} 
\begin{subtable}[t]{0.70\columnwidth} 
    \centering
    \caption{\textbf{Cosine-based TEXT-to-ECHO retrieval}}
    \label{stab:cosine_echo_recall}
    \resizebox{\columnwidth}{!}{ 
    \begin{tabular}{l?cc?c}
        \toprule
        \multirow{2}{*}{\textcolor{Purple}{ }} 
        & \multicolumn{2}{c?}{\textcolor{Maroon}{MIMIC-ECHO}}
        & \multirow{2}{*}{\textcolor{Maroon}{RSUM}} \\ 
        & R@1 & R@5 \\  
        \hline
        EchoCLIP~\cite{christensen2024visionechoclip}   & \underline{1.1} & \underline{6.4} & \underline{7.5} \\
        \hline
        PCME++~\cite{chun2023improvedpcmepp}  & 1.0 & 4.0 & 5.0 \\
        \textsc{ProbMED} (Ours)   & \textbf{1.7} & \textbf{7.8} & \textbf{9.5} \\
        \bottomrule
    \end{tabular}}
\end{subtable}
    \caption{Cross-modal retrieval performance (Recall@K) for TEXT-to-CXR, TEXT-to-ECG, and TEXT-to-ECHO retrieval tasks, \textbf{the similarity metric for all models is cosine similarity}. }
\end{table}
\subsection{Few-Shot Classification}
In ~\cref{tab:cxrfewshot},~\cref{tab:ecgfewshot},~\cref{tab:echofewshot}, we highlighted performance under a limited range of \(k\)-shot conditions to demonstrate our approach’s ability to learn effectively from scarce labeled examples. Here, we present an expanded set of few-shot experiments (including 2-, 4-, 8-, and 16-shot scenarios) for completeness and transparency. These additional results, shown in~\cref{stab:cxrfewshot_extended},~\cref{stab:ecgfewshot_extended},~\cref{stab:echofewshot_extended}, show how each model scales with varying amounts of labeled data in FS scenarioes.
\begin{table*}[ht]
  \centering 
  \resizebox{0.9\textwidth}{!}{
  \begin{tabular}{l?cccc?cccc?cccc?cccc}
    \thickhline
    \multirow{2}{*}{} 
    & \multicolumn{4}{c?}{\textcolor{Green}{Kaggle COVID}} 
    & \multicolumn{4}{c?}{\textcolor{Green}{RSNA Pneumonia}} 
    & \multicolumn{4}{c?}{\textcolor{Green}{Montgomery}} 
    & \multicolumn{4}{c}{\textcolor{Green}{CheXchoNet $\star$}} \\
    & 2S & 4S & 8S & 16S 
    & 2S & 4S & 8S & 16S
    & 2S & 4S & 8S & 16S
    & 2S & 4S & 8S & 16S \\
    \hline
    MedCLIP~\cite{wang2022medclip} 
    & 80.5 & 85.5 & 88.8 & 90.8
    & 55.7 & 58.0 & 61.8 & 65.4
    & 87.0 & 87.3 & 86.4 & 88.5
    & 52.4 & 55.8 & 59.3 & 63.9 \\
    CXR-CLIP~\cite{you2023cxrclip} 
    & 81.5 & \textbf{86.7} & \underline{89.9} & 91.6
    & 60.2 & 64.1 & 66.5 & 70.9
    & 80.1 & 85.8 & 89.1 & 91.6
    & 51.1 & 53.2 & 55.9 & 59.7 \\
    BiomedCLIP~\cite{zhang2023biomedclip} 
    & \textbf{82.8} & 86.0 & 88.8 & 89.4
    & \underline{75.6} & \underline{80.3} & \underline{83.1} & \underline{84.0}
    & 86.6 & 87.0 & 90.9 & 92.2
    & \textbf{58.8} & 59.8 & 60.9 & 61.8 \\
    CheXzero~\cite{tiu2022expertchexzero} 
    & \underline{82.2} & 82.8 & 84.7 & 88.4
    & 75.3 & 75.0 & 78.1 & 82.7
    & 85.8 & 88.5 & 90.4 & \underline{92.9}
    & 52.5 & \underline{60.3} & 60.8 & \underline{66.1} \\
    MEDBind~\cite{gao2024medbind}
    & 82.0 & 86.2 & 89.6 & \textbf{92.0}
    & 62.5 & 67.3 & 70.6 & 73.4
    & \underline{87.5} & \underline{89.9} & \underline{91.0} & 91.8
    & 56.0 & 57.6 & \underline{62.2} & 65.4 \\
     \hline

    PCME++~\cite{chun2023improvedpcmepp} 
    & 81.8 & 79.6 & 81.1 & 85.9
    & 72.6 & 73.2 & 77.0 & 79.2
    & 76.2 & 75.7 & 80.6 & 81.8
    & 51.1 & 56.8 & 61.8 & 62.7 \\
    \textsc{ProbMED} (Ours) 
    & 81.7 & \underline{86.5} & \textbf{90.4} & \underline{91.8}
    & \textbf{77.5} & \textbf{82.2} & \textbf{84.0} & \textbf{84.7}
    & \textbf{91.2} & \textbf{93.1} & \textbf{93.2} & \textbf{93.7}
    & \underline{58.0} & \textbf{63.3} & \textbf{65.3} & \textbf{68.5} \\
    %
    \thickhline
  \end{tabular}
  }
  \caption{\textbf{CXR-based few-shot extended results} (2, 4, 8, 16 shots denoted as \#S). Model performance is reported as AUROC (\%). \textbf{\textcolor{Green}{$\star$}} CheXchoNet is an \textit{emergent} dataset using CXR-to-ECHO labels.}
  \label{stab:cxrfewshot_extended}
\end{table*}
\begin{table*}[ht]
\centering
\begin{subtable}[t]{0.64\textwidth}
  \caption{\textbf{ECG-based few-shot classification.}}
  \label{stab:ecgfewshot_extended}
  \centering
  \resizebox{\linewidth}{!}{
    \begin{tabular}{l?cccc?cccc?cccc}
      \thickhline
      \multirow{2}{*}{} & \multicolumn{4}{c?}{\textcolor{Purple}{PTB-XL}} & \multicolumn{4}{c?}{\textcolor{Purple}{ICBEB}} & \multicolumn{4}{c}{\textcolor{Purple}{MUSIC $\star$}} \\
      & 2S & 4S & 8S & 16S & 2S & 4S & 8S & 16S & 2S & 4S & 8S & 16S \\
      \hline
      ECG-CLIP~\cite{gao2024medbind} & 62.7 & 67.1 & 70.0 & 71.2 & 62.7 & 69.1 & 72.0 & 74.1 & 44.7 & 48.6 & 51.1 & 51.4 \\
      MEDBind~\cite{gao2024medbind}  & 65.1 & 71.1 & 75.9 & \underline{81.8} & \underline{76.2} & \underline{81.2} & \underline{84.5} & \underline{87.8} & 50.0 & \underline{51.5} & \underline{52.5} & \underline{54.6} \\
      ECG-FM~\cite{mckeen2024ecg}   & 64.6 & 69.1 & 70.9 & 71.6 & 65.6 & 69.3 & 71.0 & 71.8 & 47.0 & 50.0 & 51.0 & 53.1 \\
      \hline
      PCME++~\cite{chun2023improvedpcmepp}   & \underline{72.6} & \underline{75.4} & \underline{77.9} & 79.9 & 65.2 & 74.1 & 79.5 & 80.5 & \underline{49.1} & 46.7 & 50.3 & 48.6 \\

      \textsc{ProbMED} (Ours) & \textbf{80.7} & \textbf{82.6} & \textbf{85.6} & \textbf{87.6} & \textbf{81.0} & \textbf{84.8} & \textbf{87.3} & \textbf{90.1} & \textbf{51.6} & \textbf{53.8} & \textbf{57.1} & \textbf{59.1} \\
      \thickhline
    \end{tabular}
  }

\end{subtable}
\hfill
\begin{subtable}[t]{0.32\textwidth}  
  \caption{\textbf{ECHO-based few-shot results.} }
  \label{stab:echofewshot_extended}
  \centering
  \resizebox{\linewidth}{!}{
    \begin{tabular}{l?cccc}
      \thickhline
      \multirow{2}{*}{} & \multicolumn{4}{c}{\textcolor{Maroon}{EchoNet-Dynamic}} \\
      & 2S & 4S & 8S & 16S \\
      \hline
      ECHO-CLIP~\cite{christensen2024visionechoclip} & \textbf{88.2} & \textbf{88.3} & \underline{93.7} & \underline{95.0} \\
      \hline
      PCME++~\cite{chun2023improvedpcmepp} & \underline{87.0} & 87.5 & 92.5 & 94.1 \\

      \textsc{ProbMED} (Ours) & 86.7 & \underline{87.7} & \textbf{95.9} & \textbf{96.2} \\
      \thickhline
    \end{tabular}
  }

\end{subtable}
\caption{Comparison of ECG-based and ECHO-based few-shot classification results. AUROC (\%) for 2-, 4-, 8-, 16-shot (\#S).\textbf{\textcolor{Purple}{$\star$}}~MUSIC is an \emph{emergent} dataset using ECG-to-ECHO labels.}
\label{stab:combined_fewshot}
\end{table*}

\begin{table}[t]
    \centering
    \label{stab:full_mimic_ckd_results}
    \resizebox{0.9\columnwidth}{!}{%
    \begin{tabular}{l?ccccc}
        \thickhline
        \multirow{2}{*}{\textbf{Method}} & \multicolumn{5}{c}{\textbf{CKD}} \\
& ZS & 2S & 4S & 8S & 16S \\
        \hline
        \multicolumn{6}{l}{\textit{CXR-only model performance}} \\
        \hline
MedCLIP~\cite{wang2022medclip}   & 61.8 & 56.7 & 59.6 & 61.6 & 62.3 \\
CXR-CLIP~\cite{you2023cxrclip}  & 73.5 & 56.5 & 59.4 & 61.5 & 62.0 \\
MEDBind~\cite{gao2024medbind}     & 71.9 & 54.0 & 54.8 & 55.5 & 57.6 \\
CheXzero~\cite{tiu2022expertchexzero} & 73.6 & 68.7 & 66.9 & 69.4 & 70.4 \\
BiomedCLIP~\cite{zhang2023biomedclip}        & 65.6 & 61.1 & 66.5 & 68.4 & 71.2 \\
PCME++~\cite{chun2023improvedpcmepp} (CXR)      & 51.0 & 55.9 & 68.6 & 69.8 & 68.8 \\
\textsc{ProbMED} (CXR)     & \underline{75.0} & \textbf{69.8} & \underline{70.6} & \underline{70.8} & \underline{76.5} \\
        \hline
        \multicolumn{6}{l}{\textit{ECG-only model performance}} \\
        \hline
ECG-CLIP~\cite{gao2024medbind}          & 54.1 & 50.9 & 58.9 & 61.0 & 66.4 \\
MEDBind~\cite{gao2024medbind}     & 61.2 & 54.9 & 66.4 & 66.9 & 67.8 \\
ECG-FM~\cite{mckeen2024ecg}     & - & 48.9 & 55.2 & 59.7 & 62.1\\
PCME++~\cite{chun2023improvedpcmepp} (ECG)            & 31.7 & 55.8 & 66.7 & 69.0 & 70.3 \\
\textsc{ProbMED} (ECG)           & 68.5 & 56.7 & 67.4 & 65.1 & 71.1 \\
        \hline
        \multicolumn{6}{l}{\textit{Using both ECG + CXR Models}} \\
        \hline
MEDBind~\cite{gao2024medbind}     & 71.5 & 54.3 & 68.4 & 69.8 & 69.8 \\
PCME++~\cite{chun2023improvedpcmepp} & 46.8 & 56.5 & 69.4 & 70.6 & 71.6 \\
\textsc{ProbMED} & \textbf{78.1} & \underline{67.5} & \textbf{71.5} & \textbf{73.4} & \textbf{76.8} \\
        \thickhline

    \end{tabular}%
    }
    \caption{\textbf{MIMIC-CKD Results}}
\end{table}
\begin{table}[t]
    \centering
    \label{stab:full_mimic_chd_results}
     \resizebox{0.9\columnwidth}{!}{
    \begin{tabular}{l?ccccc}
        \thickhline
        \multirow{2}{*}{\textbf{Method}} & \multicolumn{5}{c}{\textbf{CHD}} \\
         & ZS & 2S & 4S & 8S & 16S \\
        \hline
        \multicolumn{6}{l}{\textit{CXR-only model performance}} \\
        \hline
        MedCLIP~\cite{wang2022medclip}         & 65.4 & 64.4 & 68.2 & 73.8 & 74.7 \\
        CXR-CLIP~\cite{you2023cxrclip}        & 73.1 & 65.1 & 68.4 & 73.9 & 75.2 \\
        MEDBind~\cite{gao2024medbind} & 76.6 & 56.7 & 63.4 & 65.1 & 69.1 \\
        CheXzero~\cite{tiu2022expertchexzero}        & \underline{77.1} & 62.1 & 67.4 & 73.8 & 74.6 \\
        BiomedCLIP~\cite{zhang2023biomedclip}      & 61.8 & 61.2 & 67.0 & 69.4 & 73.7 \\
        PCME++~\cite{chun2023improvedpcmepp} (CXR)    & 51.1 & 70.1 & 72.7 & 75.6 & 76.7 \\
        \textsc{ProbMED}  (CXR)   & 77.0 & 71.0 & 71.9 & \underline{77.7} & \underline{79.8} \\
        \hline
        \multicolumn{6}{l}{\textit{ECG-only model performance}} \\
        \hline
        ECG-CLIP~\cite{gao2024medbind}        & 65.7 & 60.5 & 71.1 & 75.0 & 74.1 \\
        MEDBind~\cite{gao2024medbind} & 65.6 & 61.7 & 73.5 & 73.9 & 73.5 \\
        ECG-FM~\cite{mckeen2024ecg}     & - & 69.4 & 71.6& 72.1 & 73.9 \\
        PCME++~\cite{chun2023improvedpcmepp} (ECG)      & 40.3 & 64.7 & \underline{74.7} & 75.9 & 76.7 \\
        \textsc{ProbMED} (ECG)         & 70.3 & 64.4 & 72.2 & 73.8 & 76.7 \\
        \hline
        \multicolumn{6}{l}{\textit{Using both ECG + CXR Models}} \\
        \hline
        MEDBind~\cite{gao2024medbind}     & 75.3 & 68.3 & 71.7 & 75.9 & 78.6 \\
        PCME++~\cite{chun2023improvedpcmepp}  & 52.4 & \textbf{73.7} & \textbf{76.9} & \underline{77.7} & 78.7 \\
        \textsc{ProbMED}  & \textbf{78.4} & \underline{72.4} & 73.0 & \textbf{79.2} & \textbf{80.8} \\
        \thickhline
    \end{tabular}}
    \caption{\textbf{MIMIC-CHD Results}}
\end{table}

\subsection{CXR and ECG Combination}
Here, we provide the complete results for our CKD and CHD classification experiments, including additional performance metrics and comparisons across all evaluated methods. These extended results further validate the robustness of our findings, showing consistent gains from integrating CXR and ECG and demonstrating \textsc{ProbMED}’s advantages over competing approaches. We also include detailed ablations and per-class breakdowns to highlight the nuanced benefits of our probabilistic modeling framework in zero-shot and few-shot scenarios.
\section{Visualizations}
\subsection{Qualitative Image-Text Visualizations}
\begin{figure}[t]
    \centering
    \includegraphics[width=\columnwidth]{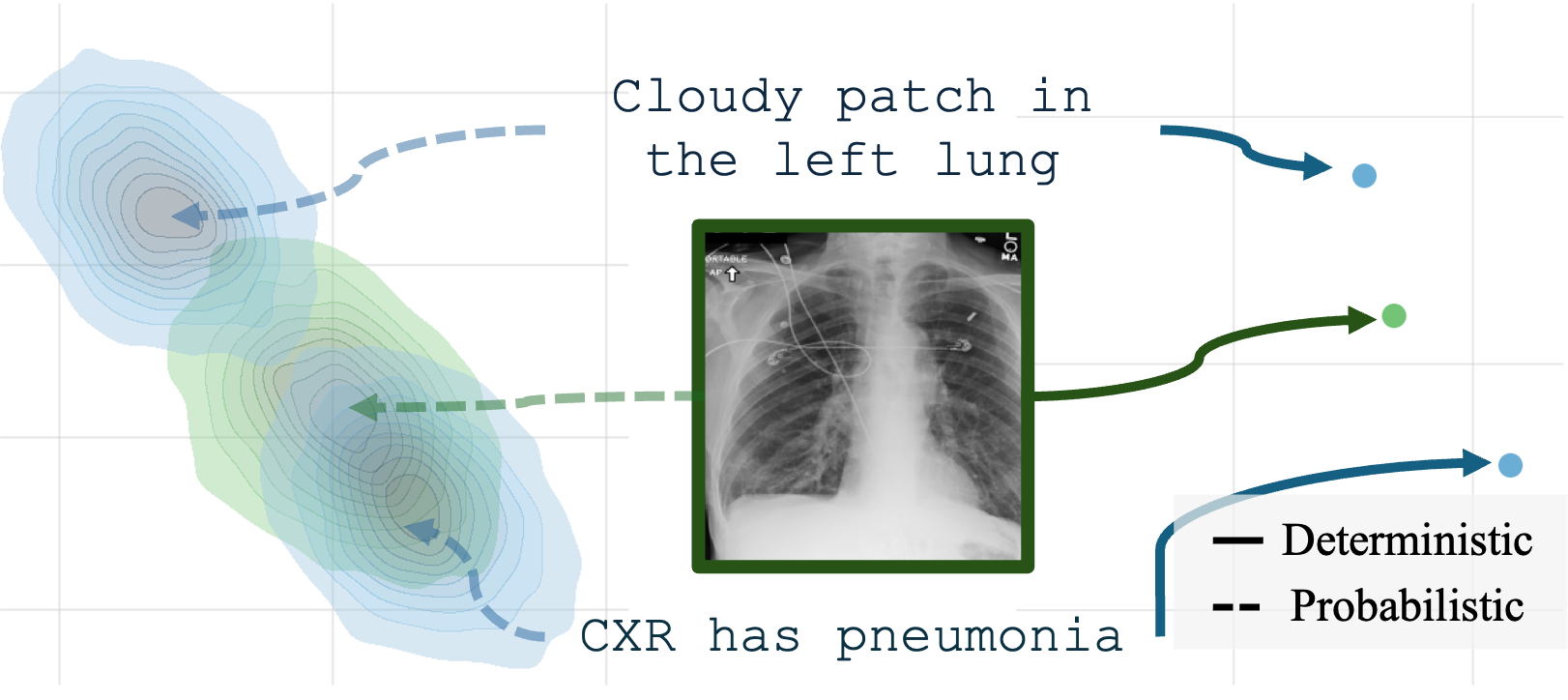}
    \caption{Qualitative visualization of \textsc{ProbMED} embeddings using PCA for dimensionality reduction. We plot a \textcolor{Green}{\textbf{CXR embedding}} (depicting pneumonia, a CXR sampled from RSNA Pneumonia dataset) alongside two distinct \textcolor{Blue}{\textbf{TXT embeddings}} of \texttt{"Cloudy patch in the left lung"} and \texttt{"CXR has pneumonia"}. In the diagram, probabilistic embeddings provide the distributions of each embedding, while the deterministic embeddings provide the limited interpretation of the ambiguity. Best viewed in color.}
    \label{sfig:model_pcavisualization}
\end{figure}

Consider a patient’s CXR showing signs of respiratory distress. Even though we can describe it with different phrases—e.g., \texttt{"CXR shows a cloudy patch in the lower lung"} or \texttt{"CXR has pneumonia"}—both statements reflect the same underlying finding. ~\cref{sfig:model_pcavisualization} illustrates this by plotting \textsc{ProbMED} embeddings for the pneumonia CXR and two text descriptions. While a purely deterministic approach (using only $\mu$) does not reveal the full similarity structure, the probabilistic embeddings (incorporating $\mu$ and $\sigma$) cluster the image and text descriptions together, highlighting the importance of modeling uncertainty in medical image–text alignment.
\begin{figure}
    \centering
    \includegraphics[width=\columnwidth]{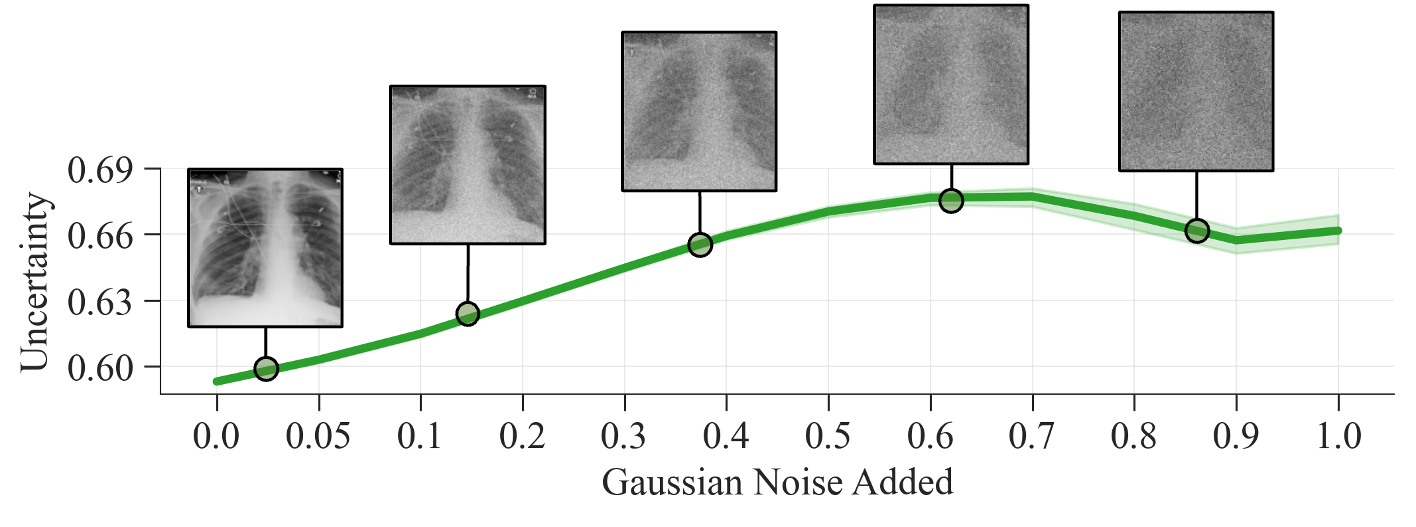}
    \caption{Qualitative visualization of \textsc{ProbMED} embeddings of a CXR with increasing Gaussian noise added. \textbf{Uncertainty} was defined and represented as the $\frac{1}{2}\exp(\cdot)$ average of the $\log(\sigma^2)$ vector across the dimension axis.}
    \label{sfig:increasingnoisevis}
\end{figure}
\subsection{Increasing Uncertainty with Noise}
~\cref{sfig:increasingnoisevis} shows how \textsc{ProbMED} responds to increasing levels of Gaussian noise injected into a CXR image. Specifically, we examine the average of the $\log(\sigma^2)$ vector across embedding dimensions—interpreted as the model’s estimated uncertainty. As noise increases, $\log(\sigma^2)$ also increases, indicating that \textsc{ProbMED} becomes more “uncertain” when the input is corrupted. This behavior makes intuitive sense as we imagine that a more pixelated CXR may encapsulate many possibilities. Ultimately,  \textsc{ProbMED}’s ability to capture variability in its latent space as it dynamically adjusts the mean and variance of its probabilistic embeddings in response to noisy inputs. Note, empirically, similar observations were seen with ECG and ECHO; we present the CXR for simplicity of visualization. 

\section{Additional Ablations}
\label{suppsec:ablations}
\subsection{Effect of Temperature on Loss}
We also varied the temperature hyperparameter within the different contrastive losses used in the study. We tried different $\tau$ parameters for consistency but kept them the same throughout the model. This was due to the associated computational cost. Herein, we show its impact on multimodal alignment. As shown in \cref{tab:temp}, smaller fixed temperatures (e.g., 0.05) increase retrieval performance on MIMIC-CXR and MIMIC-ECG relative to a trainable temperature, while a moderate temperature (0.07) achieves the highest overall RSUM. These results suggest appropriately tuning the temperature can significantly influence alignment effectiveness in contrastive learning. 

We aim to explore this phenomenon more in future studies, particularly by optimizing each temperature. However, this was not feasible at this time due to computational costs with hyperparameter tuning. 
\begin{table}[t]

\centering
\resizebox{\columnwidth}{!}{%
\begin{tabular}{c?cc?cc?cc?c}
\thickhline
\multirow{2}{*}{\textbf{$\tau$}} & \multicolumn{2}{c?}{\textbf{MIMIC-CXR}} & \multicolumn{2}{c?}{\textbf{MIMIC-ECG}} & \multicolumn{2}{c?}{\textbf{MIMIC-ECHO}} & \multirow{2}{*}{\textbf{RSUM}} \\
& \textbf{R@1} & \textbf{R@5} & \textbf{R@1} & \textbf{R@5} & \textbf{R@1} & \textbf{R@5} &  \\
\hline
\textit{Trainable} & 21.8 & 59.0 & 21.3 & 53.2 & \textbf{2.4} & \textbf{11.4} & 169.1 \\
0.05               & \textbf{49.3} & \textbf{72.0} & 46.1 & \underline{86.2} & 2.2 &  6.8 & \underline{262.6} \\
0.07               & \underline{47.9} & \underline{71.4} & \textbf{48.3} & \textbf{87.0} & \textbf{2.4} &  7.8 & \textbf{264.8} \\
0.2                & 47.6 & 71.3 & \underline{48.2} & 84.9 & \underline{2.3} &  7.6 & 261.9 \\
1                  & 47.3 & 70.9 & \underline{48.2} & 83.6 & \textbf{2.4} &  \underline{8.0} & 260.4 \\
\thickhline
\end{tabular}}
\caption{$\tau$ examination for \textsc{ProbMED} contrastive losses.}
\label{tab:temp}
\end{table}

\section{Ethical Considerations}
\textsc{ProbMED} employs a probabilistic joint embedding framework to integrate multiple medical modalities and capture clinically relevant associations. However, it is designed to uncover meaningful relationships within heterogeneous medical data. It is essential to rigorously evaluate the embeddings and their potential implications for clinical decision-making. Our framework builds upon embeddings derived from various sources, including curated clinical datasets and publicly accessible medical repositories. We advocate for continuously scrutinizing probabilistic embedding methods in medical contexts to identify and mitigate unintended associations.

\end{document}